\documentclass[sigplan,10pt, screen]{acmart}
\usepackage[labelformat=simple]{subcaption}

\usepackage{graphicx}
\usepackage{textcomp}
\usepackage{xcolor}

\usepackage{url}
\usepackage{xcolor, xspace}
\usepackage{algpseudocode, algorithm}
\usepackage{multirow}
\usepackage{float}
\usepackage{caption}
\usepackage{subcaption}
\usepackage{pifont}

\definecolor{lgray}{gray}{0.95}
\newcommand{\commentout}[1]{}
\definecolor{cadmiumgreen}{rgb}{0.0, 0.42, 0.24}

\definecolor{cgreen}{rgb}{0.0, 0.42, 0.24}
\definecolor{quitepink}{rgb}{0.858, 0.188, 0.478}

\newcommand{\cmark}{\textcolor{green}{\ding{51}}}
\newcommand{\xmark}{\textcolor{red}{\ding{55}}}

\def\BibTeX{{\rm B\kern-.05em{\sc i\kern-.025em b}\kern-.08em
    T\kern-.1667em\lower.7ex\hbox{E}\kern-.125emX}}
    
\begin{document}

\newcommand{\pname}{\texttt{Aergia}\xspace}

\newcommand{\mnist}{MNIST\xspace}
\newcommand{\fashionmnist}{FMNIST\xspace}
\newcommand{\cifarten}{Cifar-10\xspace}

\acmYear{2022}\copyrightyear{2022}
\acmConference[Middleware '22]{23rd ACM/IFIP International Middleware Conference}{November 7--11, 2022}{Quebec, QC, Canada}
\acmBooktitle{23rd ACM/IFIP International Middleware Conference (Middleware '22), November 7--11, 2022, Quebec, QC, Canada}
\acmPrice{15.00}
\acmDOI{10.1145/3528535.3565238}
\acmISBN{978-1-4503-9340-9/22/11}

\title{\pname: Leveraging Heterogeneity in Federated Learning Systems}

\author{Bart Cox}
\email{b.a.cox@tudelft.nl}
\orcid{0000-0001-5209-6161}
\affiliation{
  \institution{Delft University of Technology}
  \city{Delft}
  \state{}
  \country{Netherlands}
}

\author{Lydia Y. Chen}
\email{lydiaychen@ieee.org}
\orcid{0000-0002-4228-6735}
\affiliation{
  \institution{Delft University of Technology}
  \city{Delft}
  \state{}
  \country{Netherlands}
}

\author{Jérémie Decouchant}
\email{j.decouchant@tudelft.nl}
\orcid{0000-0001-9143-3984}
\affiliation{
  \institution{Delft University of Technology}
  \city{Delft}
  \state{}
  \country{Netherlands}
}

\begin{abstract}
Federated Learning (FL) is a popular deep learning approach that prevents centralizing large amounts of data, and instead relies on clients that update a global model using their local datasets.
Classical FL algorithms use a central federator that, for each training round, waits for all clients to send their model updates before aggregating them. In practical deployments, clients might have different computing powers and network capabilities, which might lead slow clients to become performance bottlenecks. 
Previous works have suggested to use a deadline for each learning round so that the federator ignores the late updates of slow clients, or so that clients send partially trained models before the deadline. To speed up the training process, we instead propose \pname{}, a novel approach where slow clients (i) freeze the part of their model that is the most computationally intensive to train; (ii) train the unfrozen part of their model; and (iii) offload the training of the frozen part of their model to a faster client that trains it using its own dataset.
The offloading decisions are orchestrated by the federator based on the training speed that clients report and on the similarities between their datasets, which are privately evaluated thanks to a trusted execution environment. We show through extensive experiments that \pname{} maintains high accuracy and significantly reduces the training time under heterogeneous settings by up to 
27\% and 53\% compared to  \texttt{FedAvg} and \texttt{TiFL}, respectively.
\end{abstract}


\maketitle

\section{Introduction}

Federated Learning (FL) is a decentralized and inherently privacy-preserving learning paradigm where clients collectively train a machine learning model~\cite{mcmahan2017communication,bonawitz2019towards}.
During a learning round, a federator selects a subset of the clients that return an update of the global model computed using their local dataset. Upon receiving client updates, the federator aggregates them into a global model update, which is then shared with all clients.  
Most of existing aggregation algorithms, including \texttt{FedAvg}~\cite{mcmahan2017communication} and \texttt{FedProx}~\cite{li2018federated}, are synchronous, and require the federator to collect all updates from the selected clients before moving to the next training round. 

In a practical FL system, clients might have heterogeneous computational resources and 
possess data that differ both in quantities and class distribution. 
It has been shown that both resource and data heterogeneity negatively impact the performance of a FL system~\cite{kairouz2021advances,hsieh2020non,zhao2018federated,li2018federated}. 
First, relying on a mix of weak and strong clients instead of homogeneous clients to train a model can significantly prolong the training time~\cite{chai2020tifl}. Second, a classification model trained with federated learning is less accurate when the client datasets are non independently and identically distributed (non-IID)~\cite{chen_apf}.

To mitigate the impact of weak clients, also called stragglers, the state-of-the-art methods attempt to equalize the learning speed amongst the clients by (i) partitioning them based on offline profiling~\cite{chai2020tifl}, or by (ii) dropping the updates of stragglers during the training rounds~\cite{nishio2019client, li2019smartpc}. The former approach may fall short in capturing transient heterogeneity caused by  applications possibly collocated on the clients, whereas the latter might incur a severe accuracy degradation. 
Moreover, the impact of stragglers is further aggravated when encountering non-IID data among clients. Indeed, stragglers might possess a unique dataset that is critical to the overall model accuracy. In addition, due to the privacy preserving nature of FL, it is not really possible for the federator to infer the data distribution based only on the clients model updates~\cite{li2018federated,wang2020tackling}. To limit the risk of model divergence, prior studies aggregate the non-IID client data  by adding a regularization term, like in \texttt{FedProx}~\cite{li2018federated}, or by estimating their contributions, like in \texttt{FedNova}~\cite{wang2020tackling}. However, these works implicitly assume that the client nodes are homogeneous. 

In this paper, we aim to accelerate the FL training of convolutional neural networks (CNN) in presence of stragglers and non-IID data. 
A CNN is composed of convolutional layers and fully connected layers~\cite{lecun1998gradient}, which respectively learn the representation of local data and map the extracted representation into classes. The local training of CNN entails forward and backward passes on both types of layers. 

To retain the representation of the unique  datasets of stragglers, we advocate to freeze their convolutional layers, and offload the computing and updating of the convolutional layers to strong clients.
We propose \pname\footnote{In Greek mythology, \pname{} is the personification of sloth, idleness, indolence and laziness.}, a federated learning algorithm that 
monitors the local training of selected clients and offloads part of the computing task of stragglers to strong clients that have spare and idle capacities.
\pname{} relies on a client matching algorithm that associates a straggler to a strong node based on an estimated performance gain and on the similarity between their datasets, since blindly offloading local models to nodes that have drastically different data distribution leads to weight divergence~\cite{chen_apf}.
To ensure privacy, data similarities are securely evaluated using the clients' local data distributions (i.e., the number of labels per class) in an Intel SGX enclave~\cite{costan2016intel}, which is hosted by the federator.

We implement \pname in PyTorch as a middleware running on top of Kubernetes. We evaluate \pname on three datasets, namely \mnist, \fashionmnist, and \cifarten, on different network architectures against four previous heterogeneity or non-IID aware aggregation solutions~\cite{mcmahan2017communication, wang2020tackling, li2018federated, chai2020tifl}. Our FL systems consist of a mix of 24 weak, medium and strong nodes that use a different number of CPU cores. Our evaluation results show that 
\pname achieves the highest accuracy within the lowest training time.

In a nutshell, this paper makes the following contributions:
\begin{itemize}
\item We explain how a straggler can offload the training of its model to a strong client.
\item We present an algorithm that matches the performance profile and data similarity of clients.
\item We design \pname{}\footnote{\url{https://github.com/bacox/fltk}}, a federated learning middleware for highly heterogeneous clients and non-IID data that leverages model training offloading and online client matching. \pname{} relies on a trusted execution environment (an Intel SGX enclave) so that the federator can  evaluate the similarity of client datasets
without getting access to their private class distribution.   
\item We evaluate \pname on a FL cluster built on top of Kubernetes. Our evaluation results on three datasets and several networks show that \pname effectively leverages the spare computational capacity of strong clients to achieve high accuracy in low training time.
\end{itemize}


The remainder of this paper is organized as follows. \S\ref{sec:background} provides some background on Federated Learning, data and resource heterogeneity, as well as on their impact on training time and accuracy. 
\S\ref{sec:overview} provides an overview of \pname{}, while \S\ref{sec:details} describes its algorithms and implementations details.
\S\ref{sec:evaluation} presents our performance evaluation. \S\ref{sec:relatedWork} reviews the related work. Finally, \S\ref{sec:conclusion} concludes this paper.

\section{Background and Motivation}
\label{sec:background}

In this section, we first recall necessary background on deep learning models, which are core components of the federated learning paradigm, the practical heterogeneity challenges that federated learning faces and their impact on training time and accuracy. 

\subsection{Premier on Convolutional Neural Networks}  

The state-of-the-art image classifier follows the structure of convolutional neural networks (CNN)~\cite{lecun1998gradient}, which consist of convolutional and fully connected layers. The former maps the image features into a compact representation, hence they are also referred to as feature layers. The latter are dense fully connected layers that classify the representation into one of the classes. The other difference between these two types of layers is their resource demands~\cite{cox2021masa}: convolutional layers are computationally intensive while fully connected layers are memory intensive. The training time of a client's local model can be divided into two parts: the forward pass that computes the classification outcome of images, and the backward pass that back-propagates the model weights. Consequently, the training time of a typical CNN classifier can thus further be categorized into four parts: (i) ff: forward pass on feature layers, (ii) fc: forward pass on fully connected layers, (iii) bc: backward pass on fully connected layers, and (iv) bf: backward pass on feature layers.

\begin{figure*}[htp]
     \centering
     \begin{subfigure}[t]{0.32\textwidth}
         \centering
         \includegraphics[width=\columnwidth]{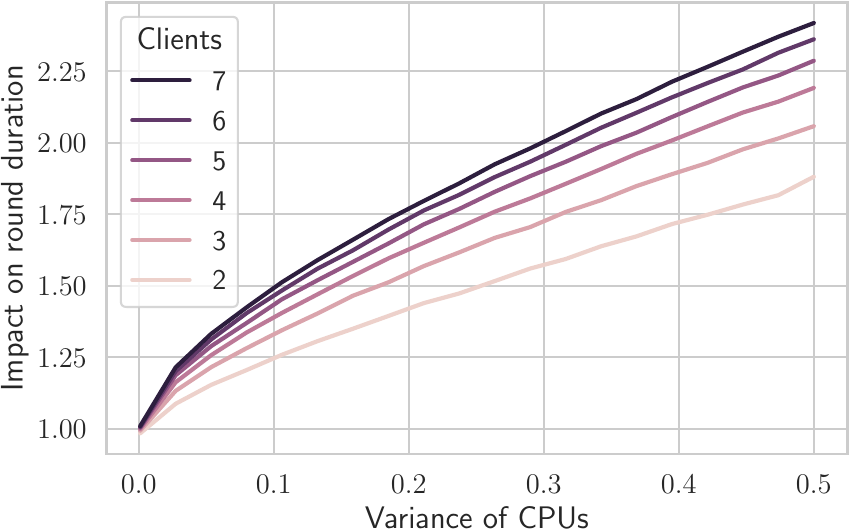}
         \caption{Impact of CPU heterogeneity among clients on training time (multiplicative factor compared to the homogeneous case). }
         \label{fig:variance_over_time}
     \end{subfigure}
     \hfill
     \begin{subfigure}[t]{0.32\textwidth}
         \centering
         \includegraphics[width=\columnwidth, ]{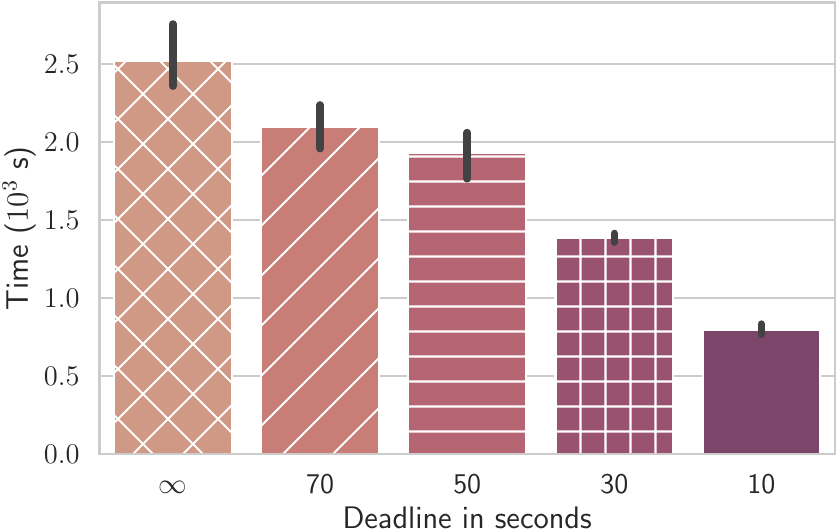}
         \caption{Total training duration in seconds without and with deadlines.}
         \label{fig:deadline_reduces_time}
     \end{subfigure}
     \hfill
     \begin{subfigure}[t]{0.32\textwidth}
         \centering
         \includegraphics[width=\columnwidth, ]{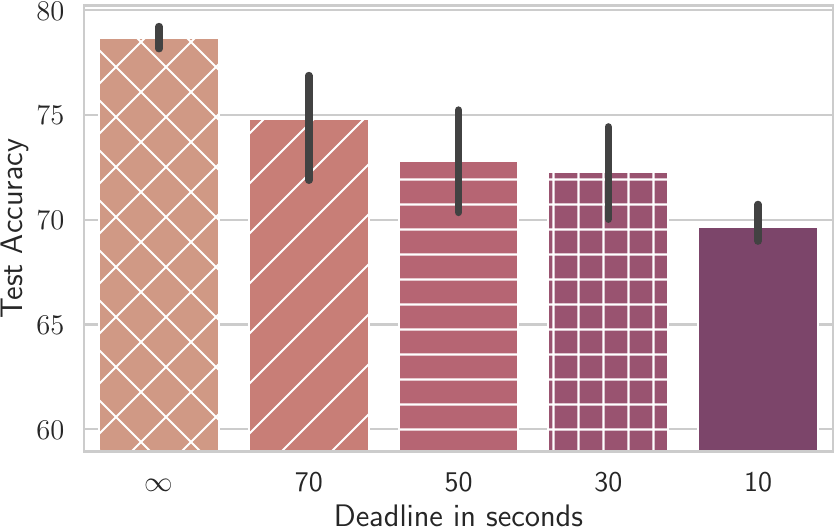}
         \caption{Accuracy in a non-IID scenario without and with deadlines.}
         \label{fig:deadline_reduces_accuracy}
     \end{subfigure}
        \caption{Heterogeneous computational powers among the clients increase the duration of the FL training process (Figure~\ref{fig:variance_over_time}). One could use deadlines so that the federator discards late updates in a round before starting the next one, which effectively reduces the training time (Figure~\ref{fig:deadline_reduces_time}). However, using deadlines badly degrades the model accuracy, in particular in non-IID settings (Figure~\ref{fig:deadline_reduces_accuracy}).}
        \label{fig:deadline_effect}
\end{figure*}

\subsection{Federated Learning}

Federated learning (FL)~\cite{kairouz2021advances,hsieh2020non,zhao2018federated,li2018federated} is an emerging decentralized learning paradigm where $K$ clients and a federator 
jointly train a machine learning model in $T$ consecutive rounds while local data stays on premise. In this paper, we specifically consider an image classification model that maps images, $x$ to one of $C$ labels, denoted by $y$,  through a function  $f(\boldsymbol{w})$, 
parameterized by weights $\boldsymbol{w}$.  Prior to the training, the federator initializes the model architecture, the objective function, the training algorithm, the training hyperparameters, 
and the aggregation protocol for the client's local update\footnote{We note that a variant of FL~\cite{ormandi2013gossip} does not rely on a federator and aggregates the model in a peer-to-peer manner. We do not consider this case here.}. We consider convolutional neural networks (CNN)~\cite{lecun1998gradient} as the classifier model. The clients train the classifier model based on their own real data, which never leaves their premises, whereas the central server iteratively aggregates and distributes models submitted from clients until reaching the global model convergence.

{\bf Local Training}. 
In each global round $t$, the clients receive the latest aggregated global model $f(w(t-1))$ from the federator and use their local data to perform local updates for $E$ epochs, e.g., stochastic gradient decent (SGD) updates~\cite{mcmahan2016federated}. The cross entropy loss function~\cite{DBLP:conf/aaai/SunCWLL16, zhang2018generalized, tang2013deep} is widely adopted for classification problems. Specifically, a client $k$ aims to find $\boldsymbol{w}_k(t)$ that  minimizes the loss function: 
$$
\min _{\boldsymbol{w}_k(t)} f_k(\boldsymbol{w}_k(t); x_k, y_k),
$$

\noindent using the $n_k$ local data points $(x_k, y_k)$, where $x_k$ is an input data, e.g., images, and $y_k$ is the class label.   
Upon finishing the local training, clients send their local model parameters, i.e., $\boldsymbol{w}_k(t)$, to the federator. 

{\bf Model aggregation}. 
After receiving all model updates from clients, the federator aggregates the clients' model parameters into the latest global model that is returned to the clients in the beginning of the next round. Specifically, at each round $t$, a subset of $K$ clients is selected to do local training and send back their latest model weights, $w_k(t)$. The aggregation algorithms differ in the frequency and weights in aggregating the local models. FedSGD~\cite{mcmahan2016federated, bonawitz2017practical} treats all local models equally, and trains the entire local data in one epoch. The gradients are sent to the federator for aggregation every epoch.  To minimize the communication and avoid the divergence of local models, \texttt{FedAvg}~\cite{mcmahan2017communication} lets local models train for multiple epochs and then aggregate the models.  Specifically, \texttt{FedAvg} calculates the global model of round $t$ as the weighted average of all $K$ local model weights:
$$\boldsymbol{w}(t)=\sum_{k=1}^{K} \frac{n_k}{\sum_{k=1}^{K} n_{k}} \boldsymbol{w}_k(t)$$

\subsection{Sources of heterogeneity}
{\bf Data heterogeneity}. Clients possess different and unique privacy-sensitive datasets. A common assumption in the prior art is that client data are identically and independently distributed, which is the so called independent and identically distributed (IID) case. Taking the image data benchmark \cifarten~\cite{krizhevsky2009learning} as an example, which contains 60,000 images from 10 classes, in the IID case each client would own an equal amount of images that would be equally distributed across classes. Recent studies point out that in practice distributed datasets are highly non-IID, and differ both in size and in distribution~\cite{wang2020tackling,arivazhagan2019federated}. For instance, it is easier to identify clients that own horse images than deer images (both are classes in \cifarten). Consequently, unique images like deer are owned by a small client subset, whereas common images like horse have a higher probability to be equally distributed across all clients.  Such non-IID data distribution, i.e., clients owning data in different quantities and distributions, have been shown to be challenging for FL and detrimental to the accuracy of the global models~\cite{kairouz2021advances, hsieh2020non, zhao2018federated}. The heterogeneity of a non-IID dataset can be captured by its Earth Mover Distance (EMD)~\cite{rubner1998metric}.
The higher the EMD of a dataset, the higher the heterogeneity of the client data distribution. To mitigate the accuracy degradation in FL due to non-IID data, related studies~\cite{li2018federated, wang2020tackling} added regularization terms in the objective function, altering the aggregation algorithms, or augmenting the dataset.

{\bf Clients resource heterogeneity}. Edge devices are highly heterogeneous in their computing and network resources~\cite{wu2019machine}. Their hardware and software stacks evolve after each generation, i.e., every 5 to 6 years. It is challenging to find an optimal (deep) learning model for a diversified set of devices. 
Due to the differences in the type and number of CPU cores and memory provisioning, the computation time of deep models on edge devices vary a lot. Moreover, the network connectivity of edge devices may often be unstable and expensive~\cite{chen_apf}. Instead of computation, communication is more a bottleneck for edge devices that have poor connectivity, questioning the effectiveness of aggregation algorithms relying on frequent communication. The heterogeneity of clients' resources essentially leads to (highly) unbalanced local training time and also long global training time, because the federator can only aggregate after receiving all the local models, including the slowest one,  during the synchronous training. To alleviate either the communication or computation bottleneck, the prior art proposes to have asynchronous communication across global round~\cite{Shiva:ICML21:communication}, i.e., the federator aggregates the local updates as soon as a new update is received. The downside of asynchronous training is the risk of slow convergence and low accuracy of the global model. Thus, the state-of-the-art mainly addresses resource heterogeneity by equalizing the communication and computation time across clients so as to maintain the synchronous training and high model accuracy. 

\subsection{Motivation: Impact of heterogeneity on training time and accuracy}
\label{sec:heterogeneity}

Resource heterogeneity in FL algorithms increases training time. \pname{} addresses resource heterogeneity thanks to its model freezing and offloading approach.
Several works also documented the negative effect of data heterogeneity among clients on FL accuracy~\cite{li2018federated, wang2020tackling, karimireddy2020scaffold}, which further increases the effect of resource heterogeneity.  \pname{} mitigates this issue by taking into account the similarities between client datasets in its offloading scheduling algorithms.  

We evaluate the influence of different degrees of CPU core heterogeneity among clients on the FL training time. We then evaluate the learning degradation that would be incurred when naively equalizing the training time of all rounds based on deadlines. For this experiment, we use the \mnist dataset, and refer the reader to Section~\ref{sec:eval_setup} for additional experimental details. 

Figure~\ref{fig:variance_over_time} shows how the overall training time of various  sizes of FL clusters increases with the variance between the clients CPUs. We consider 1 to 13 clients. We set the average computational capacity per client to be 0.5 CPU, and assign cores to clients with a variation shown in the figures. A high variance implies a high difference among slow and fast clients. We compute the overall training time as follows: we add the time required for any pre-training requirements (such as offline profiling if the algorithm demands it) to the time required for all training rounds. The duration of a training round is measured by the Federator, using its local clock, from the moment the request is sent to the clients until the last participating client responds with its results. One can see that a small perturbation in clients CPU core, i.e., standard deviation, can significantly extend the training time because the federator has to wait for the stragglers because of the synchronous training process. The delay that is added to the total training time grows with the variance of capacity as well as with the cluster size.

The state-of-the-art minimizes the impact of resource heterogeneity of clients by equalizing  their computation time through estimating clients speed and assigning different amounts of training loads. In Figures~\ref{fig:deadline_reduces_time} and~\ref{fig:deadline_reduces_accuracy}, we evaluate a naive solution - terminating the training per round according to the deadlines. Clients who cannot send back their model weights on time are not included in the aggregation, which  effectively bounds the training time but severely degrades accuracy. In contrast, the sophisticated equalization algorithms carefully factor in the trade-off between the accuracy and training time, biasing toward achieving a high accuracy. 
Motivated by this result, we 
aim to derive an algorithm that equalizes the training time within and across training rounds.

\section{Overview of \pname{}}
\label{sec:overview}

\subsection{System Model}

\begin{figure}[t]
    \center
    \includegraphics[width=.8\columnwidth]{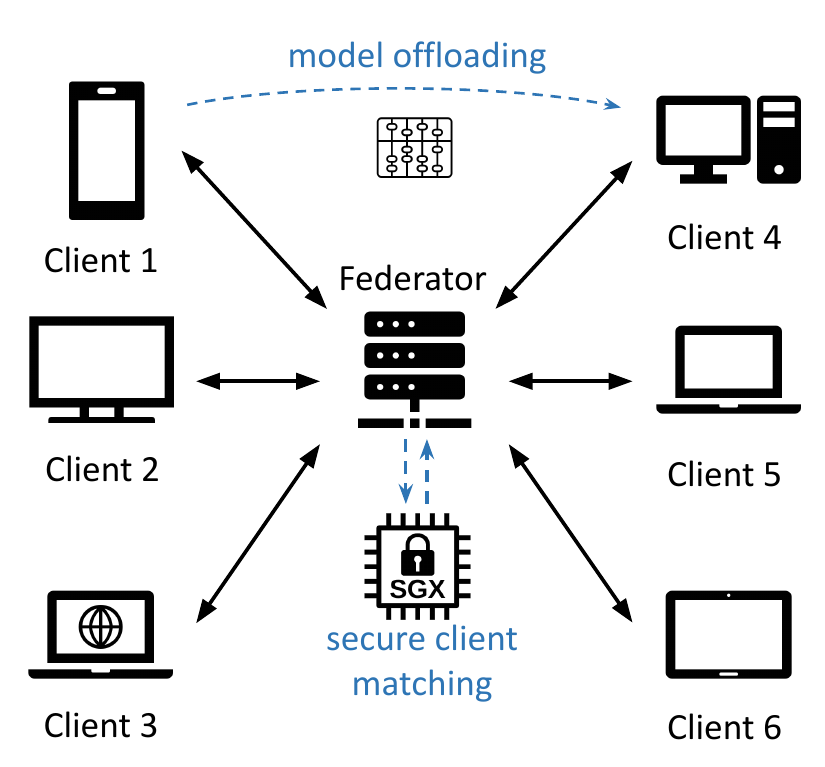}
    \caption{System architecture of \pname.}
    \label{fig:architecture}
\end{figure}

We consider a traditional federated learning architecture, which Figure~\ref{fig:architecture} illustrates, where multiple clients are connected to a central server, and are also able to directly communicate with each other. We consider heterogeneous settings in the sense that clients might have different computational powers and be interconnected by network links that differ in terms of bandwidth and latency. We consider that the computational load of each client might evolve with time, e.g., because they are running other applications in addition to the training process that we aim at optimizing. We however assume that the network properties remain stable during the training process. 
We also assume all parties to be honest. We assume that communications are asynchronous but reliable, i.e., that there is no bound on the time it takes for messages to reach their destination but all messages eventually arrive. 
Clients are equipped with local clocks so that they can measure the time they require to train their models. We do not assume these clocks to be synchronized, but they do need to have similar frequencies. We do not require the federator to make use of a clock.  
We assume the federator to be correct. It is however equipped with an Intel SGX enclave that the clients can authenticate using remote attestation~\cite{costan2016intel}, and to which they send their encrypted and privacy-sensitive dataset class distribution. The enclave provides code integrity and confidentiality of the data it manipulates. The federator relies on the datasets similarity matrix that is computed by the enclave to refine its offloading decisions and maintain high accuracy.

\pname{} allows the training process to become aware of the shortcomings and strengths of specific clients and organize their collaboration so that they rely on each other to reduce the overall training time. The central server plays a predominant role in the sense that it coordinates the overall training process and periodically identifies the clients that should offload part of their training tasks to more powerful clients through a model freezing method.  

\subsection{Leveraging the Heterogeneity of the Learning Phases}

\begin{figure}[tbp]
    \center
    \includegraphics[width=.9\columnwidth]{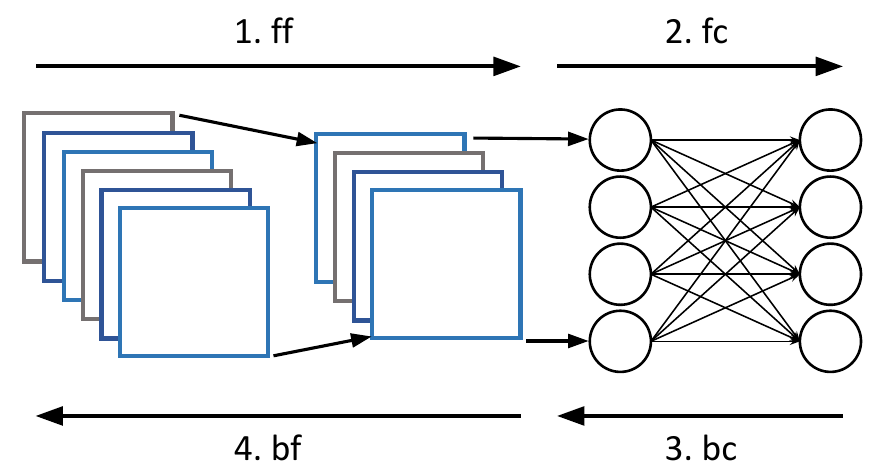}
    \caption{Model training phases during a local update: forward pass feature layers (ff), forward pass classifier layers (fc), backwards pass classifier layers (bc), and backwards pass feature layers (bf).}
    \label{fig:freezing}
\end{figure}

A CNN can be split into two sections. The first set of layers are the feature layers and the second part are the classifier layers. During training, for each minibatch the network performs a forward pass and a backward pass, respectively in the forward and in the backward propagation phases, for both sections of the layer. Figure~\ref{fig:freezing} illustrates the four phases of a training cycle: forward pass feature layers (ff), forward pass classifier layers (fc), backwards pass classifier layers (bc), and backwards pass feature layers (bf).

\begin{figure}[htp]
    \includegraphics[width=\columnwidth]{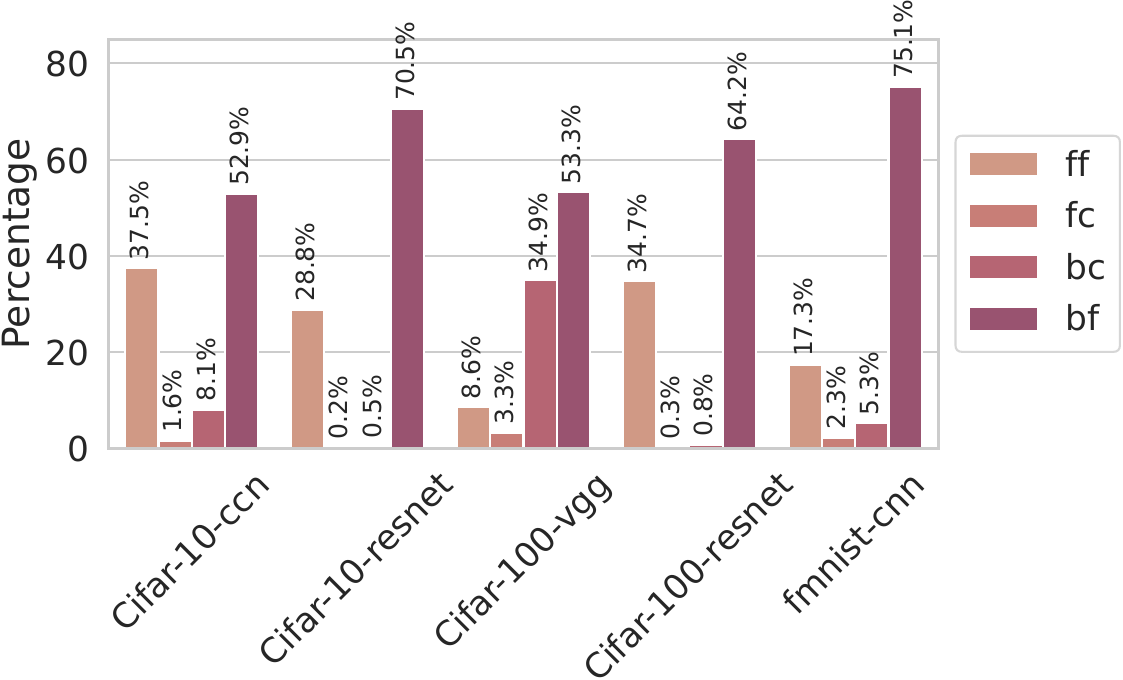}
    \caption{Profiling the different update parts of a network on several datasets shows that the execution time per phase is not uniform during a local update.}
    \label{fig:network_profiling}
\end{figure}

We have profiled the time required by each of the four phases during the network training process with several datasets. To accurately measure the computing time, our profiling time presented here is under the single client scenario.  Our results indicate that the time required to perform a cycle is not uniform across phases. Figure~\ref{fig:network_profiling} shows that the majority of the time is spent in the backwards pass of the feature layers (from 52\% to 75\%). \pname{} therefore focuses on offloading the backwards pass feature layers phase of the weak clients to stronger ones.

\pname{} leverages parameter freezing, which is commonly used for transfer learning~\cite{oquab2014learning}, personalisation tasks~\cite{arivazhagan2019federated, zhang2021parameterized} and to accelerate training~\cite{brock2017freezeout, xiao2019fast}. Freezing parameters in layers eliminates the needs to perform back propagation for the frozen parts of the network. 

\subsection{Round training with Model Freezing and Offloading}

The client selection and aggregation in \pname{} is done in the same way as in typical federated learning. For each local update, the server selects a subset of the devices and sends them the central model. 
\pname{} differs from the standard FL starting from the moment where clients start training locally. 
In the following, we describe the various tasks executed during a round, starting from the reception of the global model by the clients. Section~\ref{sec:details} provides a precise description and the implementation details of these tasks. 

\textbf{Early training and profiling.} At the beginning of a training round, clients start by executing complete batches, i.e., they use the 4 training phases, and monitor the speed at which they execute each phase. Clients then report their performance measurements, obtained from the online profiler (which directly uses the clients' clock to measure processing times), to the central server, 
and continue fully training their model while waiting for the server's instructions. 

\textbf{Centralized scheduling.} When it has received the performance measurements of all clients, the server identifies the clients that would slow down the entire training process by sending their updated model later than others, and computes a schedule where slow clients offload part of their training process to more powerful and data compatible clients. This computation takes the clients dataset similarities into account by leveraging an Intel SGX enclave for privacy. The server indicates to the slow clients that they should offload their model to a stronger client, whose IP address is specified and who will train part of their model for them. The server also informs the stronger clients that should train the model of weaker clients. Messages from the server detail the global training round number so that clients can ignore late messages.  

\textbf{Model freezing, offloading and training.} Clients are informed of the offloading they might have to execute according to the schedule the server computed, and the time at which all clients are expected to send their updated model to the server. Weak clients that need to offload their models freeze the first layers of their model, send their model to a stronger client and continue updating their model with a lighter procedure. Strong clients that are executing offloaded tasks return their own fully trained model and the (possibly partially trained) model they might have been updating for a weaker node at the time of the deadline.

\textbf{Model aggregation.} Upon receiving all model updates, the server recombines the models of clients that offloaded their training process: the first layers are received from the stronger client to which their training was offloaded, and the remaining layers are received directly from the weaker client. Based on the reconstructed models and based on the models that were fully trained by clients, the server uses the classical FL averaging method to compute the next global model, and start the next global update.  

\section{System Details of \pname{} }
\label{sec:details}

In this section, we detail the key components of \pname{}. 
We first describe \pname{}'s model freezing and offloading algorithm.
We then detail how the federator relies on profiling results communicated by clients to schedule the models freezing and offloading that should happen during a global round. We finally explain how the federator leverages an  SGX enclave to compute the clients dataset similarities to refine its schedule. 

\subsection{Model Freezing and Offloading}

\begin{figure}[htbp]
    \includegraphics[width=\columnwidth]{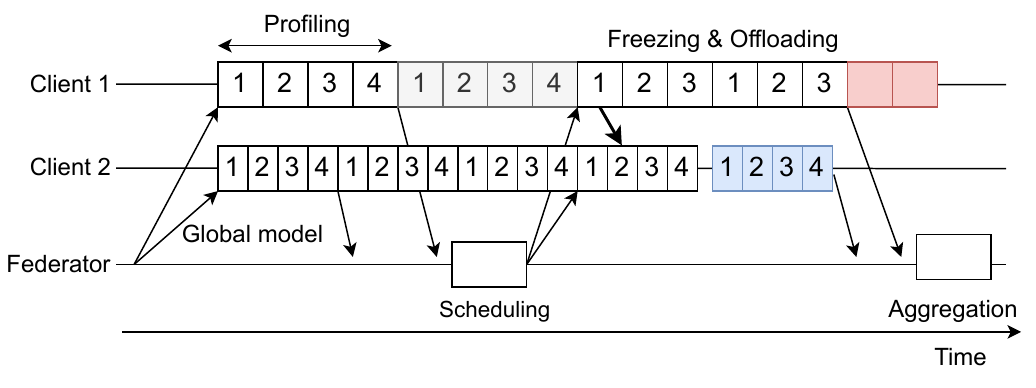}
    \caption{Illustration of model freezing and offloading. Client 1 freezes and offloads its model to client 2 to accelerate the training process. }
    \label{fig:offloadingScheduling}
\end{figure}

Figure~\ref{fig:offloadingScheduling} illustrates the potential impact of task offloading on the duration of a learning round with heterogeneous clients, and allows us to explain how model freezing and offloading intervene in \pname{}. At the beginning of a round, the server sends the current global model to the clients, after which they compute their local model using their own datasets. In this example, clients 3 and 4 are more powerful than clients 1 and 2, which impacts the time it takes them to execute a training phase (numbered from 1 to 4). In this example, the clients profile their performance over a number of local batch updates (\pname uses 100 batches), and report their performance metrics to the server afterwards. After having collected all performance measurements, the server determines which clients should freeze the second part of their model and offload its training to other clients. The server then informs the clients to freeze their model and to offload the training of their last layers to a selected strong client. Symmetrically, the federator also communicates with the strong clients about the offloading decision so that they are aware of the identity of the weak client whose model they should train, and know the number of local training batches they should use to train it.  While they are waiting for the server's scheduling decision, clients keep training their models using all four phases. In Figure~\ref{fig:offloadingScheduling}, clients 1 and 2 are informed that they should respectively freeze and offload their models to clients 3 and 4. When clients 3 and 4  finish executing their local updates (here after 3 batches), they train the offloaded models they have received on their local datasets. All clients eventually return the local models they train (offloaded or not) to the server, which finally aggregates them. In this example, clients 1 and 2 execute their rounds faster than they would have without offloading because they save the execution of two phases. Note that scheduling decisions are cryptographically signed by the federator for authenticity, and that they contain a monotonically increasing sequence number so that they cannot be replayed and so that messages sent by the federator that arrive late (i.e., in the next round) are ignored.   

\subsection{Online Profiling}
Minimizing the maximum training time of all clients during a local update is equivalent to  minimizing the variance between the client training times. To reach this goal, \pname{} gathers information about the training speed of each client over the first local updates of a global training round. This information is initially not available to the federator, and might also evolve over time as clients might dedicate a various proportion of their computing power to the learning task. During the profiling phase, clients keep track of the execution time of the forward- and backward propagation of each layer. These execution times are then communicated to the server and allows it to obtain a clear picture of the current computational power of the clients, and identify the clients that are slowing down the whole learning process. The profiling phase should include sufficiently enough batches to report accurate numbers, but not too many so that there is still headroom to optimize the remaining part of the learning process (whose entire duration depends on client dataset sizes). \pname{} profiles the training process during 100 local batch updates out of 1600 local batch updates\footnote{The number of local upates used for the profiler depends on the size of the dataset and the batch size. For our experiments we use 100 local batch updates to limit the profiling overhead and still achieve a good profiling accuracy.}.

\subsection{Centralized Scheduling}
\label{sec:centralizedScheduling}

\begin{algorithm}[htbp]
    \caption{Scheduling models freezing and offloading with $m$ clients} 
    \label{alg:scheduling}
    \begin{algorithmic}[1]
    \State \textbf{Inputs:}
        \State \hspace*{2em} $t_{j,\{1,2,3\}}$, $t_{j,4}$: training times of client $j$ on tasks $(1,2,3)$ and $4$, for $1 \le j \le m$.
        \State \hspace*{2em} $ru_{j}$: remaining local updates of client $j$.
        \State \hspace*{2em} $S_{i,j}$: Similarity values between clients $i$ and $j$, describing pair wise data similarity. 
        \State \hspace*{2em} $f$: Similarity factor. 
    \State
    
    \State \textbf{Outputs:}
        \State \hspace*{2em} $deadlines$: array of times at which a client must send the partially or completely trained models to the server  
        \State \hspace*{2em} $sending$: an array that details the stronger client to which a client should offload its model to (possibly None).
    \State
    
    \State \textbf{Algorithm:} 

        \State \hspace*{0em} $mct = \frac{1}{|M|}\sum^{M}_{m=1} ru_m * (t_{m,\{1,2,3\}} + t_{m,4})$
        \State \hspace*{0em} $sending = \{x | \forall x \in M, ru_x * (t_{x,\{1,2,3\}} + t_{x,4}) > mct\}$ \label{alg:line:create_sending}
        \State \hspace*{0em} $receiving = \{y | \forall y \in M, ru_y * (t_{y,\{1,2,3\}} + t_{y,4}) \leq mct\}$\label{alg:line:create_receiving}
        
        \State \hspace*{0em} $sort\_ascending(sending)$
        \State \hspace*{0em} $sort\_descending(receiving)$
        \State \hspace*{0em} $deadlines = []$

        \State \hspace*{0em} \textbf{for} each $c \in sending$:
            \State \hspace*{1em} $selected\_client = None$
            \State \hspace*{1em} $offloading\_cost = \infty$
            \State \hspace*{1em} $op = 0$
            \State \hspace*{1em} \textbf{for} each $k \in receiving$:
                \State \hspace*{2em} $ct, d = calc\_op(t_{c\{1,2,3,4\}}, t_{k\{1,2,3,4\}}, t_{k\{4\}}, ru_c, ru_k)$\label{alg:line:call_opt_time}
                \State \hspace*{2em} $cost_{temp} = ct * (1 + \log(S_{c, k} * f + 1))$ \label{alg:line:cost_function} 
                \State \hspace*{2em} \textbf{if} $cost_{temp} < offloading\_cost$:
                \State \hspace*{3em} $offloading\_cost$ = $cost_{temp}$
                \State \hspace*{3em} $selected\_client$ = k
                \State \hspace*{3em} $op = d$
            \State \hspace*{1em}  $receiving =  receiving - selected\_client$ \label{alg:line:remove_receiving}
            \State \hspace*{1em} $deadlines[c] = (selected\_client, op)$
            \State \hspace*{1em} \textbf{if} $receiving = \emptyset$:
                \State \hspace*{2em} \textbf{break}
        \State \hspace*{0em} \textbf{return} $(deadlines, sending)$ 
            
    \end{algorithmic}
    \end{algorithm}

\begin{algorithm}[htbp]
    \caption{$calc\_op()$: Calculating the optimal offloading point ($op$) between two nodes} 
    \label{alg:scheduling_opt_time}
    \begin{algorithmic}[1]
    \State \textbf{Inputs:}
        \State \hspace*{0em} $t_a$: training time for client $a$.
        \State \hspace*{0em} $t_B$: training time for client $b$.
        \State \hspace*{0em} $x_b$: training time of only conv layer for client $b$.
        \State \hspace*{0em} $r_a$: remaining local updates of client $a$.
        \State \hspace*{0em} $r_b$: remaining local updates of client $b$.
    \State
    
    \State \textbf{Outputs:}
        \State \hspace*{0em} $ct$: Estimated duration between two nodes.
        \State \hspace*{0em} $d$: Number of local updates to be executed before offloading to the other node.
    \State
    
    \State \textbf{opt\_time($t_a, t_b, x_b, r_a, r_b$):} 
        \State \hspace*{0em} $ct = \infty$
        \State \hspace*{0em} \textbf{for} $d \in 1\ldots \min(r_a, r_b):$
            \State \hspace*{1em} $current\_ct = \max((r_a - d) * t_a + d * x_b, (r_b - d) * t_b)$
            \State \hspace*{1em} \textbf{if} $current\_ct > ct:$
                \State \hspace*{2em} \textbf{return} $ct$, $d$
            \State \hspace*{1em} $ct= current\_ct$ 
        \State \hspace*{0em} \textbf{return} $ct$, $d$
        
    \end{algorithmic}
    \end{algorithm}

The federator in \pname{} executes a scheduling algorithm that considers the training process of each client as two consecutive tasks that can potentially be executed on different clients. The first task of each client is always executed locally while the second task can be offloaded to another client. 
Based on the performance metrics gathered by the profiler, 
the offloading scheduling algorithm decides whether the second task of each client should be offloaded to another client. In Figure~\ref{fig:offloadingScheduling}, clients 1 and 2 respectively offload the second task of their training process to clients 3 and 4. Clients 3 and 4 execute these offloaded tasks as soon as they receive them and are done computing their own model updates. Overall, thanks to the offloaded tasks, the training time of the overall round can be reduced.

When it has received the performance measurements of all clients, the server identifies the clients that would slow down the entire training process by sending their updated model later than others, and computes a schedule where slow clients offload part of their training process to more powerful and data compatible clients. The server directly informs the slow clients that should offload their training process to another client.

The main objective of the server is to minimize computational variance by correctly matching a weak client with a strong client. The central server is during the local training of the clients informed about the training performance of the clients. The clients gather these performance estimates with the online profiler while training locally on the data. Note that the profile has a very low overhead as it only represents 0.58\% of the total training time.  
Using the performance indicators, the central server matches underperforming clients with strong clients following Algorithm~\ref{alg:scheduling}. 
We estimate a \textit{mean compute time} for each round based on the performance data of each client that is active in the current round. The central server uses the \textit{mean compute time} ($mct$) as a target time for the offloading schedule. Using $mct$ and the performance indicators of each client, the weak and strong clients are identified (Line \ref{alg:line:create_sending} and \ref{alg:line:create_receiving}). Pairs of weak and strong clients are created such that the compute time of the weak clients approaches the initial \textit{mean compute time}. Similarly, the additional offloading work of the strong client should be bounded by  \textit{mean compute time}, i.e., its original compute time and offloading time should not exceed the \textit{mean compute time}.
Clients are sorted based on their expected training duration. The set of clients who are {not able to meet the imposed \textit{mean compute time}} are the weak clients. For these clients, a suitable strong client needs to be identified. The scheduling algorithm matches clients starting by the weakest ones because the global training time in a round is determined by the weakest client. For a possible offloading option, the algorithm evaluates the optimal offloading time (Line \ref{alg:line:call_opt_time}). The optimal offloading time is calculated using the performance indicators of two nodes as can be seen in Algorithm \ref{alg:scheduling_opt_time}. 
A strong client can only be used one time per round for offloading. Once a strong client is assigned to a weak client, the strong client is removed from the list of possible receiving nodes (Line \ref{alg:line:remove_receiving}).
One can modify Algorithms~\ref{alg:scheduling} and~\ref{alg:scheduling_opt_time} to support heterogeneous network transmission latencies and bandwidths, which is straightforward and has been omitted for space reasons.     

Globally minimizing the time required to train all client models during a FL round can be seen as a $Qm | r_j | C_{max}$ job scheduling problem, which can be solved using dynamic programming~\cite{lawler89sequencing}. Our algorithm is a variant of the greedy longest-processing-time-first (LPT) algorithm that distributes the training tasks over the clients, which has been demonstrated to produce schedules that are close to optimal~\cite{graham1969bounds}. In addition, LPT scales linearly with the number of clients, and therefore does not significantly slow down the federator. 

{\bf{Training procedure}}.
The scheduling algorithm is used by the federator to orchestrate the training process. The actions of the Federator during a training round consist of the following steps. First, the client selection, in the same manner as with \texttt{FedAvg}, is performed. The selected clients are by the Federator instructed to start training for $T$ rounds with the online profiler active. Each client after $P$ number of local batch updates informs the federator about the performance metrics that are gathered with the profiler. Simultaneously, the clients stop the online profiler and continue training until further instruction.
With all the performance metrics that the federator receives, an offloading schedule is created using Algorithm \ref{alg:scheduling}. The clients are informed by the federator about the offloading schedule.
The model aggregation is performed when the federator has received a training result from all selected clients.
This process is repeated for $T$ rounds.

\subsection{Refining Schedules with Data Heterogeneity}

So far, the scheduling algorithm we have presented does not take into account the fact that clients might possess non-IID local datasets. 
In practice, offloading the model of a client to a stronger client with a vastly different dataset would be detrimental to the training accuracy.  The central server therefore also takes into account the similarity of two datasets to compute the model freezing and offloading that should take place during a round.
The similarity between datasets is computed by the federator using the data distribution over the classes. We implemented this process using a trusted execution environment, namely Intel SGX, but it could also be implemented using cryptographic methods such as homomorphic encryption. We demonstrate that using dataset similarities allows \pname{} to improve its model accuracy and leave such privacy-preserving extensions to future work.  
The local data distribution of a client is privacy sensitive, therefore, we calculate the similarity matrix $S$ inside a trusted execution environment (i.e., an SGX enclave) of the federator node. Clients send an encrypted vector that contains the number of labels per class to the enclave. The trusted execution environment computes the pair-wise similarity between pairs of clients based on the EMD metric. The result is a similarity matrix $S$, a $m \times m$ triangular matrix where $m$ is the number of clients. The similarity matrix $S$ enables \pname{} to improve the offloading algorithm without publicly sharing sensitive information.

The cost function on line \ref{alg:line:cost_function} of Algorithm \ref{alg:scheduling} takes the data similarity of two clients into account. The similarity matrix $S$ is calculated before the training starts, and it contains the pair wise similarity of the local data using the EMD metric. 
Variable $f$ on line \ref{alg:line:cost_function} of Algorithm \ref{alg:scheduling} is used to control the effect of the inter client similarity. With $f=0$, the scheduling only takes the performance metrics into account and ignores the local data similarities, as discussed in section~\ref{sec:centralizedScheduling}. As soon as $f > 0$, the inter-client data similarity is taken into consideration while creating the schedule. A larger $f$ increases the influence of data similarity when determining the offloading decisions.
Using this simple objective function, the scheduling algorithm associates each weak client to a suitable stronger client that will partially train its model on a dataset that is sufficiently similar to its own.

\section{Evaluation}
\label{sec:evaluation}

\begin{figure*}[htp]
    \centering
    \begin{subfigure}[t]{1\textwidth}
      \centering
          \includegraphics[height=.55cm]{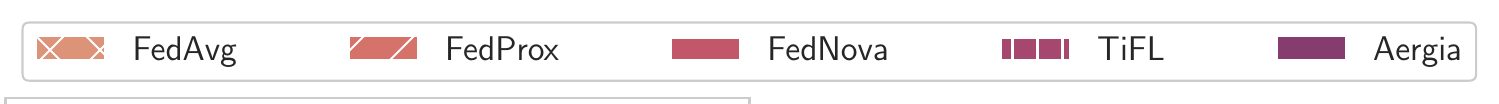}
      \end{subfigure}
    \begin{subfigure}[t]{0.16\textwidth}
        \centering
        \includegraphics[width=\columnwidth, height=2.5cm]{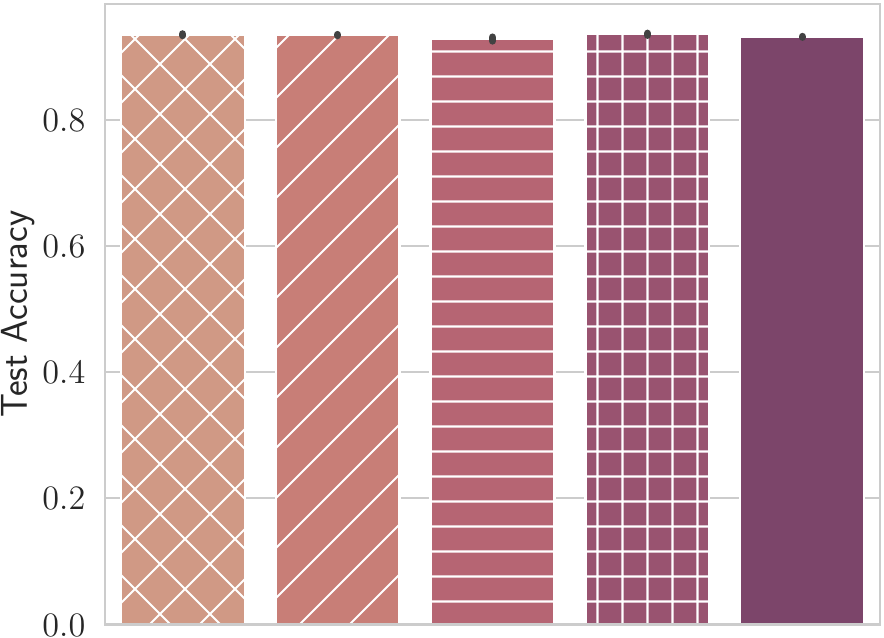}
        \caption{Accuracy \mnist}
        \label{fig:eval_mnist_acc_iid}
    \end{subfigure}
    \hfill
    \begin{subfigure}[t]{0.16\textwidth}
        \centering
        \includegraphics[width=\columnwidth, height=2.5cm]{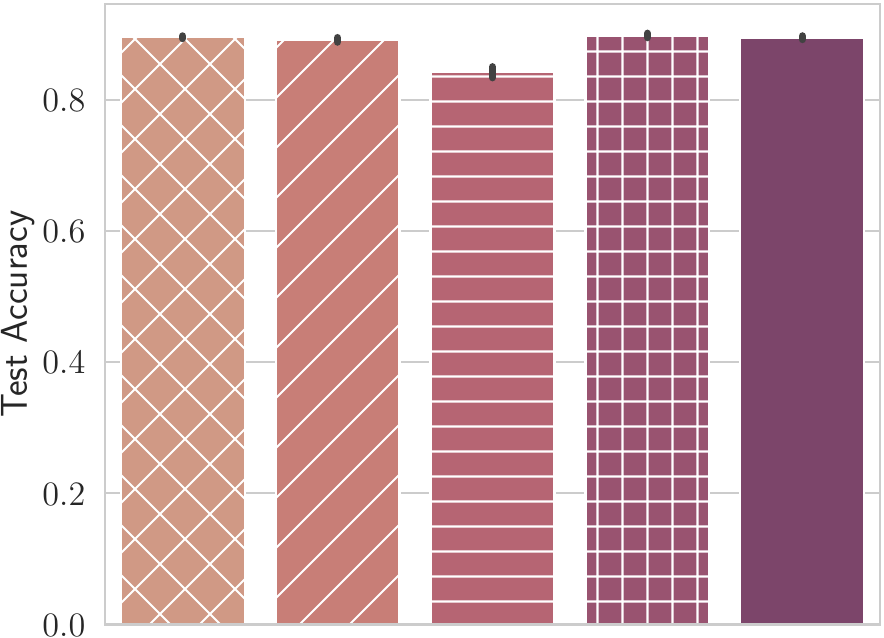}
        \caption{Accuracy \fashionmnist}
        \label{fig:eval_fashion_mnist_acc_iid}
    \end{subfigure}
    \hfill
    \begin{subfigure}[t]{0.16\textwidth}
        \centering
        \includegraphics[width=\columnwidth, height=2.5cm]{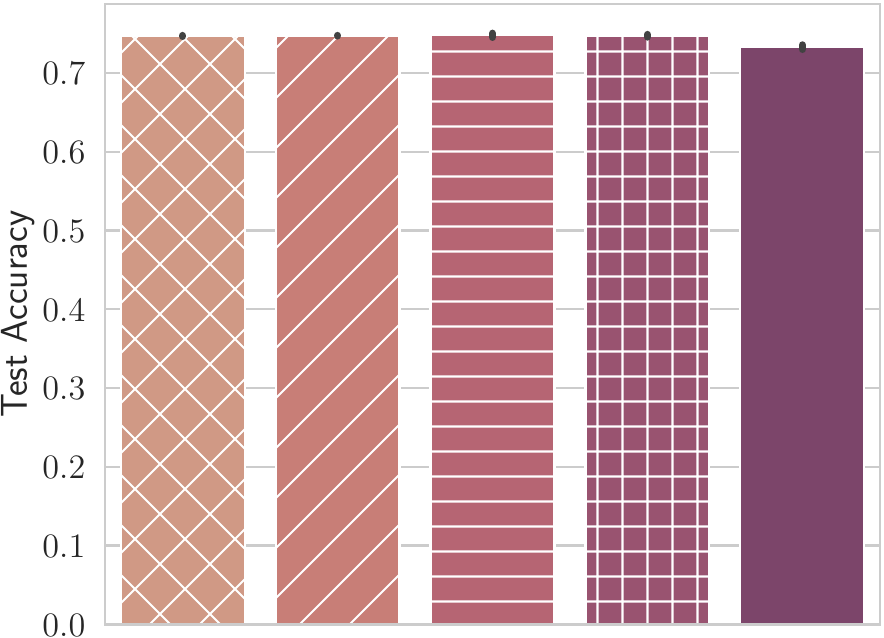}
        \caption{Accuracy \cifarten}
        \label{fig:eval_cifar10_acc_iid}
    \end{subfigure}
    \hfill
    \begin{subfigure}[t]{0.16\textwidth}
        \centering
        \includegraphics[width=\columnwidth, height=2.5cm]{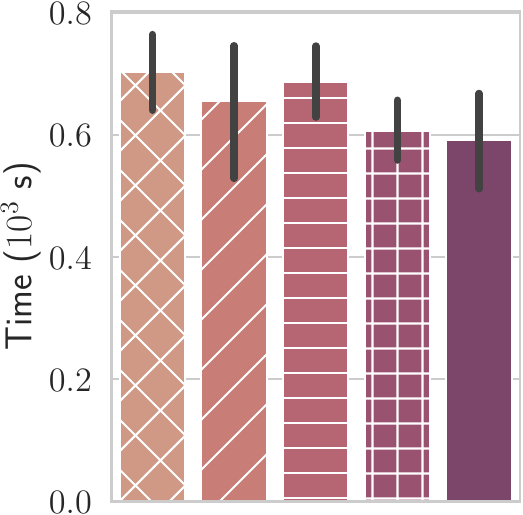}
        \caption{Time \mnist}
        \label{fig:eval_mnist_time_iid}
    \end{subfigure}
    \hfill
    \begin{subfigure}[t]{0.16\textwidth}
        \centering
        \includegraphics[width=\columnwidth, height=2.5cm]{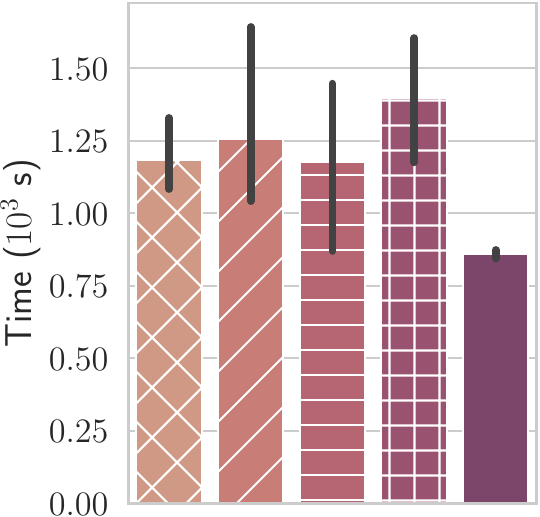}
        \caption{Time \fashionmnist}
        \label{fig:eval_fashion_mnist_time_iid}
    \end{subfigure}
    \begin{subfigure}[t]{0.16\textwidth}
        \centering
        \includegraphics[width=\columnwidth, height=2.5cm]{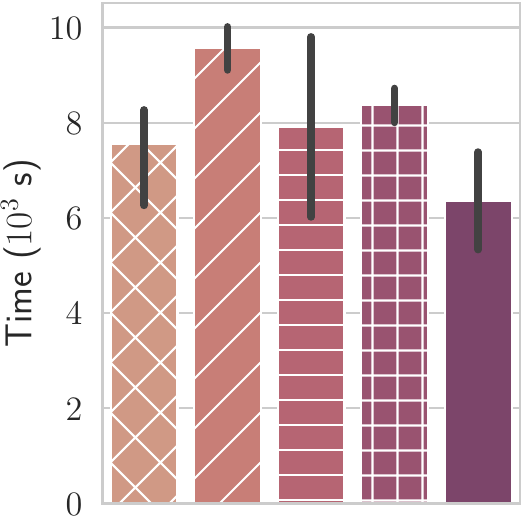}
        \caption{Time \cifarten}
        \label{fig:eval_cifar10_time_iid}
    \end{subfigure}
      \caption{IID Data set. The time reported in subfigures \ref{fig:eval_mnist_time_iid}, \ref{fig:eval_fashion_mnist_time_iid}, \ref{fig:eval_cifar10_time_iid} is the training time in seconds used to complete $100$ communication rounds.}
      \label{fig:evaluations_iid}
\end{figure*}

\begin{figure*}[htp]
    \centering
    \begin{subfigure}[t]{1\textwidth}
      \centering
          \includegraphics[height=.55cm]{figures/FIG8.pdf}
      \end{subfigure}
    \begin{subfigure}[t]{0.16\textwidth}
        \centering
        \includegraphics[width=\columnwidth, height=2.5cm]{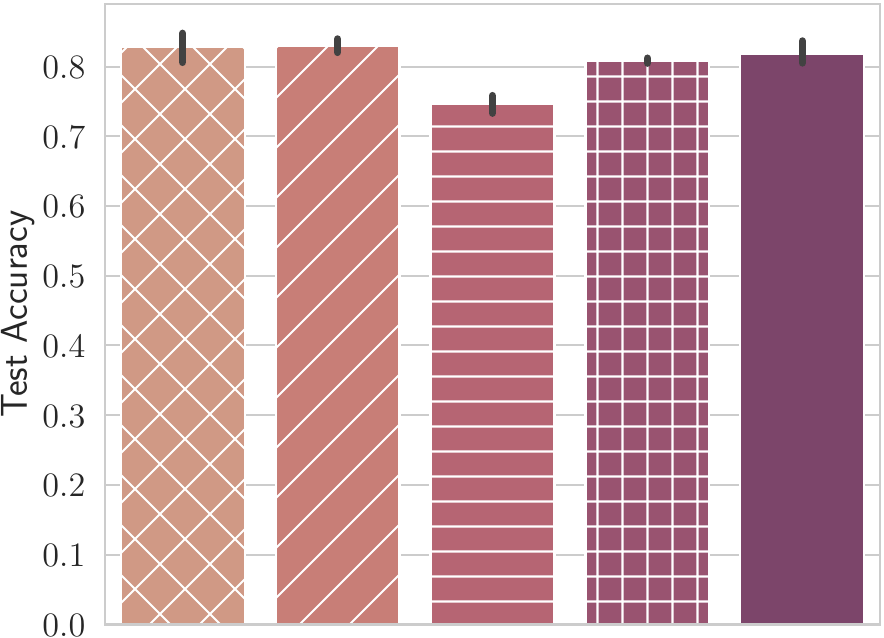}
        \caption{Accuracy \mnist}
        \label{fig:eval_mnist_acc_non_iid}
    \end{subfigure}
    \hfill
    \begin{subfigure}[t]{0.16\textwidth}
        \centering
        \includegraphics[width=\columnwidth, height=2.5cm]{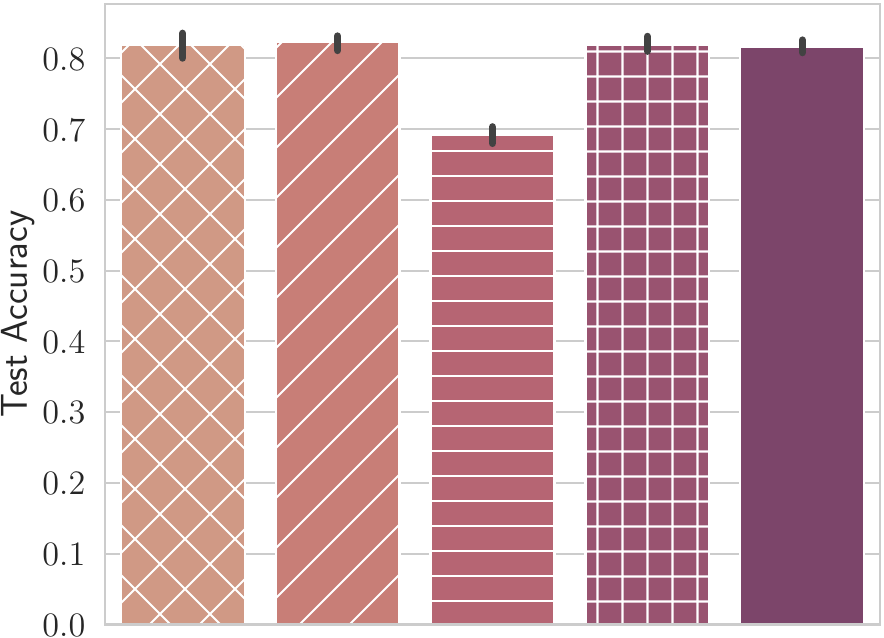}
        \caption{Accuracy \fashionmnist}
        \label{fig:eval_fashion_mnist_acc_non_iid}
    \end{subfigure}
    \hfill
    \begin{subfigure}[t]{0.16\textwidth}
        \centering
        \includegraphics[width=\columnwidth, height=2.5cm]{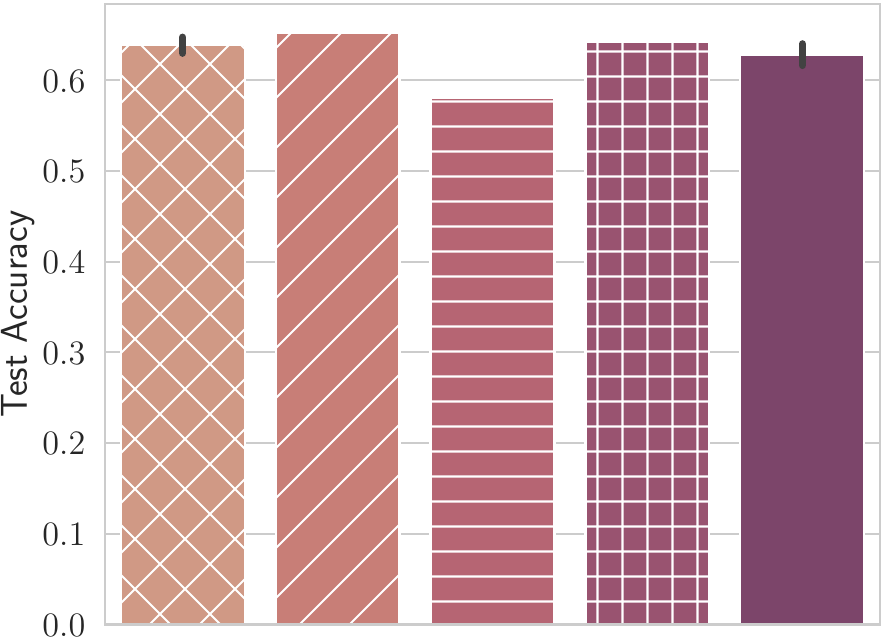}
        \caption{Accuracy \cifarten}
        \label{fig:eval_cifar10_acc_non_iid}
    \end{subfigure}
    \hfill
    \begin{subfigure}[t]{0.16\textwidth}
        \centering
        \includegraphics[width=\columnwidth, height=2.5cm]{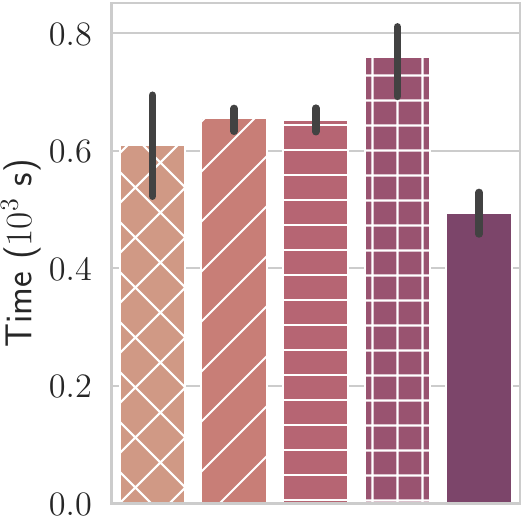}
        \caption{Time \mnist}
        \label{fig:eval_mnist_time_non_iid}
    \end{subfigure}
    \hfill
    \begin{subfigure}[t]{0.16\textwidth}
        \centering
        \includegraphics[width=\columnwidth, height=2.5cm]{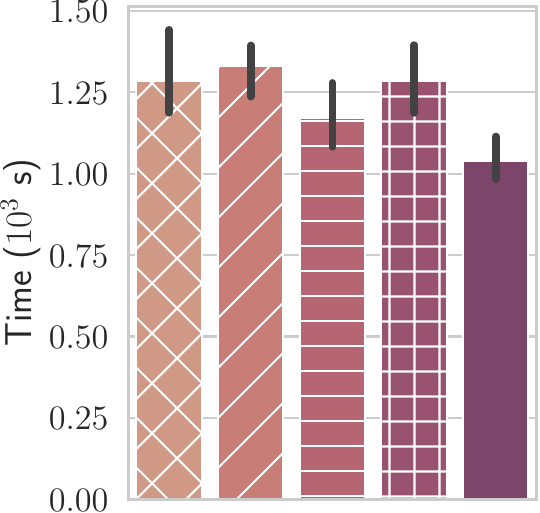}
        \caption{Time \fashionmnist}
        \label{fig:eval_fashion_mnist_time_non_iid}
    \end{subfigure}
    \begin{subfigure}[t]{0.16\textwidth}
        \centering
        \includegraphics[width=\columnwidth, height=2.5cm]{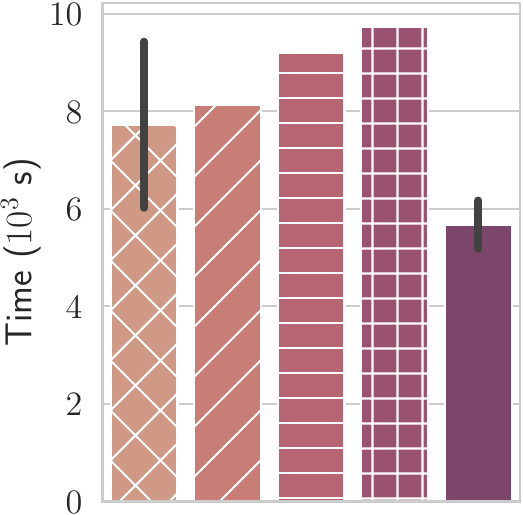}
        \caption{Time \cifarten}
        \label{fig:eval_cifar10_time_non_iid}
    \end{subfigure}
    \caption{Non-IID Data set. The time reported in subfigures \ref{fig:eval_mnist_time_non_iid}, \ref{fig:eval_fashion_mnist_time_non_iid}, \ref{fig:eval_cifar10_time_non_iid} is the training time in seconds used to complete $100$ communication rounds.}
      \label{fig:evaluations_non_iid}
\end{figure*}

In this section, we summarize the evaluation results of \pname{} on a testbed, where heterogeneous clients own diversified data sets and  learn the CNN classifier in a federated manner. We compare the overall training time and accuracy against four state of the art solutions on three datasets. We also conduct a sensitivity analysis to demonstrate the effectiveness of \pname{} under different degrees of non-IIDness. 

\subsection{Evaluation Setup}
\label{sec:eval_setup}

The central federator performs the default FL tasks (client selection and model aggregation) and also computes the client matching for the offloading of model training. During the local training, the client sends the federator performance metrics of the local training. With this information, the federator can spot stragglers that slow down the training process. Using these performance metrics, the federator matches stragglers with powerful clients in order to offload computational tasks. The matched clients communicate directly and transfer the tasks to each other. 

\textbf{Baselines.} We compare \pname{} to four baselines:\\ \texttt{FedAvg}~\cite{mcmahan2017communication}, \texttt{FedNova}~\cite{wang2020tackling}, \texttt{FedProx}~\cite{li2018federated}, and \texttt{TiFL}~\cite{chai2020tifl}. \\
\texttt{FedAvg}, \texttt{FedNova}, and \texttt{FedProx} implicitly assume that client nodes are homogeneous and that communication is synchronous. \texttt{FedProx} minimizes the amount of drift a client can obtain during local training to improve the convergence when learning over non-IID data.
\texttt{FedNova} normalizes the client updates at the aggregation stage.
\texttt{TiFL} handles stragglers on a global level by grouping the clients in tiers based on their computation speed.

\textbf{Testbed.}
The evaluations are performed on a testbed aimed to run FL systems in a distributed environment. The testbed is implemented in Python and built on top of PyTorch. Each node is fully isolated from each other and can only communicate through messages. All communication is asynchronous and peer to peer and based on RPC. This means that nodes can message each other directly without relying on the Federator as a relay (Most FL systems assume a star network). In our evaluation we assume a fully connected network. The testbed allows all nodes in the system to run independent and isolated from each other. 
We use an Aurora R13 desktop with a 5.20 GHz i9 CPU with 24 cores and 64 GB RAM. We use up to one client per CPU core. 
We implemented the dataset similarity computation that runs in an Intel SGX enclave in C++ and used Graphene~\cite{tsai2017graphene}.

\textbf{Datasets.}
We evaluate \pname{} using three datasets: \mnist, \fashionmnist and \cifarten.
The \mnist and \fashionmnist datasets contain 60,000 training images and 10,000 test images with a dimension of 28x28 pixels. The \cifarten dataset has a total of 60,000 images of 32x32 pixels split over 50,000 training images and 10,000 test images.

\textbf{Networks}
There are three networks used to evaluate the datasets.
For the \mnist dataset we use a three layer CNN with two convolutional layers and a single fully connected layer.
For \fashionmnist we use the same model as used to evaluate \mnist.
Lastly, an eight layer CNN is used to evaluate \cifarten consisting of six convolutional layers and two fully connected layers.

\textbf{Heterogeneous Resource Setup.}
In the real world, machines are rarely equal in computing power. Therefore, using large discrete steps, such as 0.25, 0.5, 0.75 and 1.0 CPU, to define resource heterogeneity between nodes is not realistic. We use a total of 24 cores to evaluate all the algorithms we consider. To reproduce a real-world scenario, the CPU speed of each client is selected uniformly at random as a fraction that ranges between 0.1 and 1.0 of the original CPU speed. We use Docker containers to isolate the nodes on different cores and control their CPU speed. This allows for an realistic approximation of the aforementioned scenario. 

\textbf{Heterogeneous Data Distribution.}
The clients that participate in the Federated Learning process may have non-uniform data distributions. To evaluate performance under non-IID heterogeneity, clients sample 3 classes out of the 10 available.
Clients are then biased towards their local dataset during training. Local client datasets are disjoint, i.e., they do not share images. This approach is used for all three datasets when evaluating the non-IID scenario.

\subsection{Accuracy and training times}
We divide our evaluation into two scenarios: IID and non-IID on three datasets. We report the accuracy and training time in Figures~\ref{fig:evaluations_iid}-\ref{fig:evaluations_non_iid}.

\textbf{IID:}~
When the local training data has a IID distribution, the effects of the resource heterogeneity are negligible. In figure \ref{fig:eval_fashion_mnist_acc_iid}-a we compare the accuracy of \texttt{FedAvg}, \texttt{FedNova}, \texttt{FedProx}, \texttt{TiFL}, and \pname on the \fashionmnist dataset. The accuracy after 100 rounds is comparable among the aforementioned algorithms. In contrary to the accuracy values, the results of the real-time duration show that there is a noticeable difference between \pname and the rest. On average, \pname is able to do the same amount of training as \texttt{FedAvg} and \texttt{TiFL} in 27\% and 45 \% less time respectively.

{\textbf{Non-IID:}~ The impact of non-IID data is intensified by the system heterogeneity and leads to a significant longer training times. This effect can be seen in Figure \ref{fig:eval_cifar10_time_non_iid}. \pname is able to reduce the training time per round and thereby reduce the overall global training time by 27\% compared to \texttt{FedAvg} and 53\% compared to \texttt{TiFL}. The same trend is visible in Figures \ref{fig:eval_mnist_time_non_iid} and \ref{fig:eval_fashion_mnist_time_non_iid}. The accuracy reached in non-IID scenarios is comparable between the \pname and the other algorithms with the exception of \texttt{FedNova}. Figures \ref{fig:eval_fashion_mnist_time_non_iid} and \ref{fig:eval_cifar10_time_non_iid} show that \texttt{TiFL} is unable to prevent the slowdown of the FL system, in contrary to what is reported in \cite{chai2020tifl}. We think this is caused by the higher variation of CPU power in this scenario. \texttt{TiFL} is in this scenario not able to reduce the variance in CPU using their tier solution due to a higher intra-tier CPU variance. Overall we observe that \pname is able to reduce the training time up to 53\% while achieving the comparable accuracy as the state of the art non-IID aware aggregation algorithms.}

\begin{figure}[htp]
    \includegraphics[width=\columnwidth]{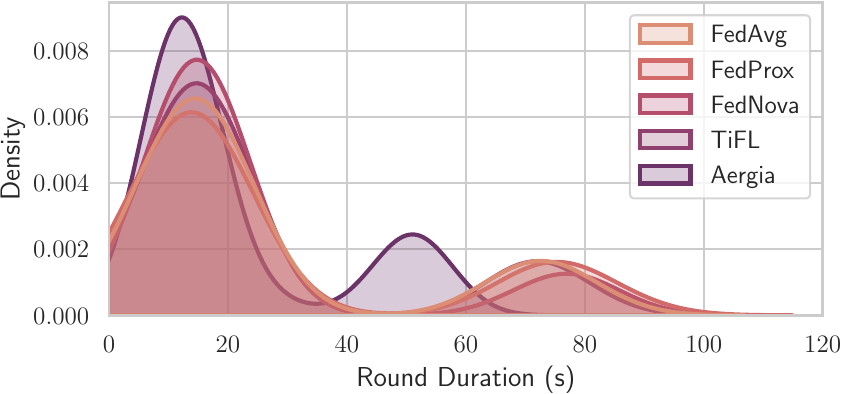}
    \caption{Density of the duration of rounds. }
    \label{fig:round_duration}
\end{figure}

To better understand the performance advantages of\\ \pname{}, we zoom into the detailed performance of \fashionmnist, i.e., the time distribution per training round.  Figure~\ref{fig:round_duration} shows the density of the rounds duration during the training process for \pname{} and the baselines we consider. \pname{}'s distribution is shifted to the left compared to all baselines, which indicates its ability to minimize the duration of rounds during training.


\begin{figure*}[htp]
     \centering
     \begin{subfigure}[t]{0.45\textwidth}
         \centering
    \includegraphics[width=\columnwidth]{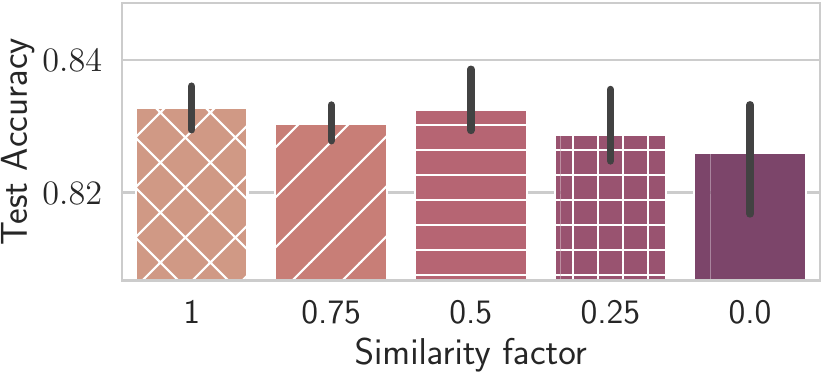}
    \caption{Impact of the client similarity factor (parameter f in Algorithm~\ref{alg:scheduling}) on accuracy when offloading. }
    \label{fig:accuracy_similarity_matching}
     \end{subfigure}
     \hfill
     \begin{subfigure}[t]{0.45\textwidth}
         \includegraphics[width=\columnwidth]{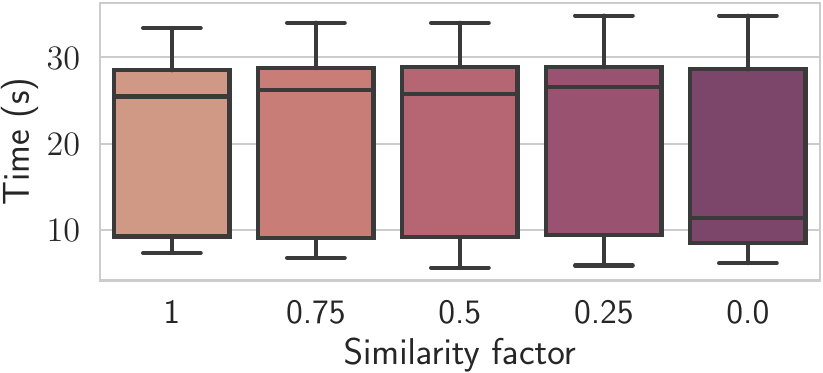}
    \caption{Impact of the client similarity factor (parameter f in Algorithm~\ref{alg:scheduling}) on the training time.}
    \label{fig:round_duration_similarity_matching}
     \end{subfigure}
        \caption{Impact of similarity factor on accuracy and training time.}
        \label{fig:duration_similarity_matching}
\end{figure*}
\subsection{Rounds duration and impact of the similarity factor}
\pname uses the similarity factor $f$ in Line \ref{alg:line:cost_function} in Algorithm \ref{alg:scheduling} to control the impact of the inter-client data similarity. A value of 0 means that the inter-client similarity has no effect on the scheduling while $f > 0$ increases the impact of similarity in Algorithm \ref{alg:scheduling}. A higher value of $f$ restricts the number of favourable strong offloading options. Figure \ref{fig:round_duration_similarity_matching} shows that the average round time decreases when similarity is not taken into account. A low values of $f$ has impact on the global model accuracy as can be seen in Figure \ref{fig:accuracy_similarity_matching}. A positive value of $f$ increases the global model accuracy in \pname.


We evaluate  learning \fashionmnist on 24 clients, 3 of which are selected in each round.
Figures~\ref{fig:accuracy_similarity_matching} and~\ref{fig:round_duration_similarity_matching} report the impact of the similarity factor parameter on accuracy and on the training time, respectively. 
With a similarity factor of 0, the scheduling algorithm ignores the inter-client similarity and uses performance indicators only. With a positive similarity factor, the inter-client similarity has more impact on the scheduling decisions.
Increasing the similarity factor can restrict the available offloading options and thereby increases the average round time.

\subsection{Impact of the Degree of Non-IIDness}
\begin{figure}[htbp]
            \includegraphics[width=\columnwidth]{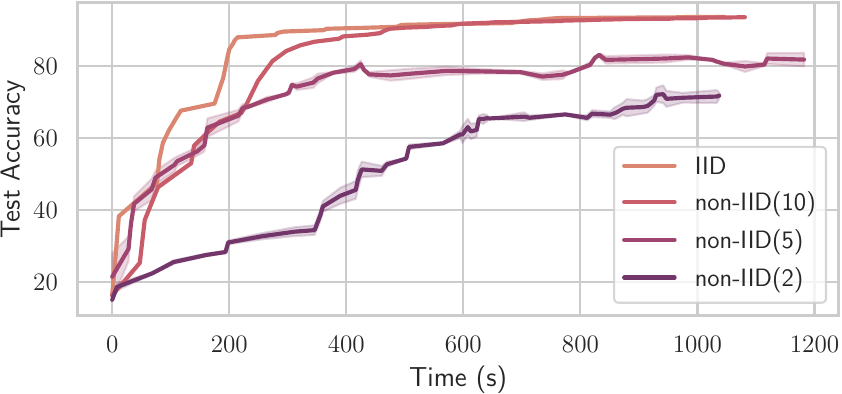}
    \caption{Test accuracy depending on the level of dataset non-IIDness measured during the training process. The level of non-IID limits the number of sampled classes in a clients' local training data. The levels of non-IIDness follow the setup used in~\cite{chai2020tifl}.}
    \label{fig:level_non-iidness}
\end{figure}

The degree of data non-IIDness has an impact on the convergence speed during training. We control the number of classes a client can own as a measure to evaluate and create non-IID data. The lower the number of sampled classes the higher the degree of non-IID in the clients local data. For instance, non-IID(2) indicates each client only has 2 classes of data points out of 10 available classes. In Figure \ref{fig:level_non-iidness} we show the effect of the degree of non-IIDness on the \pname{}'s algorithm. All variations train for the same amount of rounds. The difference in completion time is low but the difference in accuracy is apparent. A high degree of non-IID data results in a decrease of model accuracy. This is comparable to previously reported results~\cite{chai2020tifl}.

\textbf{Profiler}
The profiling time of the online profiler impacts the error in the performance indicators used for scheduling. A longer profiling time offers better performance indicators, but a longer profiling time reduces the training speed and increases the training time. The profiler has a negligible overhead of $0.22\% \pm 0.09$ on average for all models.

\section{Related Work}
\label{sec:relatedWork}

\begin{table}[t]
    \begin{tabular}{p{1.8cm}|p{1.6cm}|p{1.8cm}|p{1.5cm}}
           & \textbf{Data heterogeneity aware}  & \textbf{Resource heterogeneity aware} & \textbf{Minimize training time} \\        
        \hline
        FedAvg~\cite{mcmahan2017communication}   &  -   & - & \xmark \\ 
        FedProx~\cite{li2018federated}           &  +   & -  & \xmark\\ 
        FedNova~\cite{wang2020tackling}          &  +    & -  & \xmark \\ 
        TiFL~\cite{chai2020tifl}                 &  +    & +  & \cmark \\ \bottomrule 
        \pname                                  & ++    & ++ & \cmark \\ \bottomrule
    \end{tabular}
    \caption{FL solutions for heterogeneous settings.}
    \label{tab:comparison-solutions}
\end{table}

Devices may differ in computational and communication capabilities due to hardware (CPU, memory) and network connectivity.
Challenges such as straggler mitigation, client drift, and the effect of non-IID data are exacerbated by the aforementioned system-level characteristics. 
\texttt{FedProx}~\cite{li2018federated} limits how far the local model of the client is allowed to drift from the global model.
The objective function is altered with a parameter $\mu$ to penalize drift from the global model. Their theory shows that choosing $\mu > 0$ should improve convergence when learning over non-IID data, but practice shows that this is not the case for non-IID label distributions. The solution of Li et al.~\cite{li2018federated} does not account for stragglers.

\texttt{FedNova}~\cite{wang2020tackling} limits the effect of heterogeneous clients in the system by altering the global aggregation rule.
Clients that perform more steps return a larger update and thus more significantly impact the global model, even when the datasets of all clients have the
same size. Although Wang et al.~\cite{wang2020tackling} mitigate the effect that heterogeneous nodes have on the model accuracy, it is not able to remove the under representation of stragglers in federated learning systems.
Table~\ref{tab:comparison-solutions} summarizes our analysis of the existing works that consider heterogeneous client resources or datasets, and shows that \pname{} is the only algorithm that minimizes the actual training time, thanks to its model freezing and offloading mechanisms, and maintains high accuracy even in presence of non-IID datasets, thanks to its similarity computation that is executed in an SGX enclave.

\subsection{Task Offloading}

Dong et al.~\cite{DBLP:conf/mass/DongZ0G20} use an edge server equipped with a Trusted Execution Environment (TEE) to offload tasks of edge devices.
Because of the limited capabilities of the TEE, this method is only tested on  the \mnist dataset on a very small network. Besides that, the data is also randomly (but equally) distributed over the clients. The main goal of this work is to mitigate the straggler problem by utilizing the more powerful edge servers for task offloading.
EAFL~\cite{ji2021computation} offloads federated learning tasks from edge clients to nearby edge servers. To lighten to computational burden of weak clients, EAFL partially offloads data to an edge server, while \pname{} offloads models, which is more privacy preserving. 
FedAdapt~\cite{wu2021fedadapt} leverages split learning to allow an IoT device to train its model with the help of the central server by exchanging partially trained models during each round. In comparison, we leverage model freezing and allow clients to offload training tasks to each other, which is more scalable, and consider the data heterogeneity issue, without which accuracy would be decreased.  

\subsection{Federated Learning with deadlines}
One solution to mitigate stragglers is for the federated learning system to impose a deadline for the clients to submit their locally trained models. While this reduces the overall idle in the whole system, it often diminishes the contribution of the straggler by dropping late submission. Li et al.~\cite{li2019smartpc} use the imposed deadline to tune the local settings such as CPU frequency on each client to meet the deadline with the least amount of energy consumption. This approach lowers the amount of dropped clients, however the effect of non-IID data is not considered. Nishio and Yonetani~\cite{nishio2019client} use resource aware client selection (FedCS) to limit the amount of participating stragglers. They show that this solution works mainly works in an IID data settings in large networks. When dealing with non-IID data, models trained with FedCS incur a significant drop in accuracy.

\subsection{Stragglers}
Different techniques can be used to minimize the impact of stragglers. \texttt{TiFL}~\cite{chai2020tifl} groups clients in tiers based on their performance characteristics. Each round a different tier is chosen for client selection. This reduces the variance between training times within each round. 

Sageflow~\cite{park2021sageflow} uses periodic aggregation rounds to limit the decrease of training time per round caused by stragglers. Late contributions send in by stragglers are incorporated into the global model corrected with a staleness factor. The downside of Sageflow is that the method uses public data distributed across the clients.

Lee et al.~\cite{lee2021adaptive} use adaptive deadlines to accelerate the convergence to the global model. By calculating a deadline per round based on the participating devices, the mean idle time per device is minimized. This solution still can cause stragglers to be dropped by the system, since the deadline is calculated based on the fastest participating client plus the tolerated waiting time. The authors note that their solution has no guarantees when applied on non-IID data.

\subsection{Parameter freezing}
Parameter freezing is a well known technique in transfer learning that is used to fine tune the model. An alternative use for Parameter Freezing is to speed up the training process. Chen et al.~\cite{chen_apf} limit the communication between the federated and edge device by freezing parameters in the early stage of the training process.
However, challenges are that the local parameters excluded from global
synchronization may diverge on different clients, and meanwhile
some parameters may only temporally stabilize. Adaptive Parameter
Freezing (APF) proposes to fix (freeze) the non-synchronized stable
parameters in intermittent periods.

In order to reduce the communication cost, parameter freezing can be used. Brock et al.~\cite{brock2017freezeout} use parameter freezing to speed up the computations by slowly freezing out the first few layers. This avoids the cost of computing the gradients and speeds up convergence. This approach is coarse and degrades the accuracy of model in spite of a speedup in training. Chen et al.~\cite{chen_apf} use parameter freezing to lower the communication cost when sending the model of weights. By freezing stale parameters the number model weights that are shared with the server are reduced. While this significantly reduces data transfers, it does not alleviate the computational burden of stragglers.

Veit and Belongie~\cite{veit2018convolutional} use partial execution of networks to speed up the model execution during inference. 
By excluding some layers in the network from execution based on the input image, the inference execution time is reduced while retaining accuracy. 
Similarly, Wu et al.~\cite{wu2018blockdrop} use layer dropping in residual networks to create dynamic execution path to reduce the execution time during inference.

\section{Conclusion}
\label{sec:conclusion}

In practical settings, Federated Learning (FL) is largely impacted by the heterogeneity of clients, as each training round needs to wait for slow clients, which are also called stragglers. In this work, we have proposed \pname{} a novel FL algorithm that leverages model freezing and offloading. In this protocol, slow clients freeze the first layers of their model and offload it to a faster client that trains these layers using its own dataset. \pname{} uses a simple, yet fast and scalable, scheduling algorithm to regularly organize the offloading of models between clients. The central server uses its own dataset to decide which clients have compatible datasets to maintain high accuracy when organizing model freezes and offloadings. Our experiments on three datasets demonstrate that \pname{} reduces the overall training time by up to 53\% and maintains high accuracy.

\bibliographystyle{ACM-Reference-Format}
\bibliography{MiddleWare-FL-Task-Offloading}


\begin{thebibliography}{40}


\ifx \showCODEN    \undefined \def \showCODEN     #1{\unskip}     \fi
\ifx \showDOI      \undefined \def \showDOI       #1{#1}\fi
\ifx \showISBNx    \undefined \def \showISBNx     #1{\unskip}     \fi
\ifx \showISBNxiii \undefined \def \showISBNxiii  #1{\unskip}     \fi
\ifx \showISSN     \undefined \def \showISSN      #1{\unskip}     \fi
\ifx \showLCCN     \undefined \def \showLCCN      #1{\unskip}     \fi
\ifx \shownote     \undefined \def \shownote      #1{#1}          \fi
\ifx \showarticletitle \undefined \def \showarticletitle #1{#1}   \fi
\ifx \showURL      \undefined \def \showURL       {\relax}        \fi
\providecommand\bibfield[2]{#2}
\providecommand\bibinfo[2]{#2}
\providecommand\natexlab[1]{#1}
\providecommand\showeprint[2][]{arXiv:#2}

\bibitem[\protect\citeauthoryear{Arivazhagan, Aggarwal, Singh, and
  Choudhary}{Arivazhagan et~al\mbox{.}}{2019}]%
        {arivazhagan2019federated}
\bibfield{author}{\bibinfo{person}{Manoj~Ghuhan Arivazhagan},
  \bibinfo{person}{Vinay Aggarwal}, \bibinfo{person}{Aaditya~Kumar Singh},
  {and} \bibinfo{person}{Sunav Choudhary}.} \bibinfo{year}{2019}\natexlab{}.
\newblock \showarticletitle{Federated Learning with Personalization Layers}.
\newblock \bibinfo{journal}{\emph{CoRR}}  \bibinfo{volume}{abs/1912.00818}
  (\bibinfo{year}{2019}).
\newblock
\showeprint[arXiv]{1912.00818}
\urldef\tempurl%
\url{http://arxiv.org/abs/1912.00818}
\showURL{%
\tempurl}


\bibitem[\protect\citeauthoryear{Avdiukhin and Kasiviswanathan}{Avdiukhin and
  Kasiviswanathan}{2021}]%
        {Shiva:ICML21:communication}
\bibfield{author}{\bibinfo{person}{Dmitrii Avdiukhin} {and}
  \bibinfo{person}{Shiva Kasiviswanathan}.} \bibinfo{year}{2021}\natexlab{}.
\newblock \showarticletitle{Federated Learning under Arbitrary Communication
  Patterns}. In \bibinfo{booktitle}{\emph{Proceedings of the 38th International
  Conference on Machine Learning, {ICML} 2021, 18-24 July 2021, Virtual Event}}
  (July 18 - 24, 2021) \emph{(\bibinfo{series}{Proceedings of Machine Learning
  Research})}, \bibfield{editor}{\bibinfo{person}{Marina Meila} {and}
  \bibinfo{person}{Tong Zhang}} (Eds.), Vol.~\bibinfo{volume}{139}.
  \bibinfo{publisher}{PMLR}, \bibinfo{pages}{425--435}.
\newblock
\urldef\tempurl%
\url{https://proceedings.mlr.press/v139/avdiukhin21a.html}
\showURL{%
\tempurl}


\bibitem[\protect\citeauthoryear{Bonawitz, Eichner, Grieskamp, Huba, Ingerman,
  Ivanov, Kiddon, Kone{\v{c}}n{\'y}, Mazzocchi, McMahan, Overveldt, Petrou,
  Ramage, and Roselander}{Bonawitz et~al\mbox{.}}{2019}]%
        {bonawitz2019towards}
\bibfield{author}{\bibinfo{person}{Kallista~A. Bonawitz},
  \bibinfo{person}{Hubert Eichner}, \bibinfo{person}{Wolfgang Grieskamp},
  \bibinfo{person}{Dzmitry Huba}, \bibinfo{person}{Alex Ingerman},
  \bibinfo{person}{Vladimir Ivanov}, \bibinfo{person}{Chlo{\'{e}} Kiddon},
  \bibinfo{person}{Jakub Kone{\v{c}}n{\'y}}, \bibinfo{person}{Stefano
  Mazzocchi}, \bibinfo{person}{Brendan McMahan}, \bibinfo{person}{Timon~Van
  Overveldt}, \bibinfo{person}{David Petrou}, \bibinfo{person}{Daniel Ramage},
  {and} \bibinfo{person}{Jason Roselander}.} \bibinfo{year}{2019}\natexlab{}.
\newblock \showarticletitle{Towards Federated Learning at Scale: System
  Design}. In \bibinfo{booktitle}{\emph{Proceedings of Machine Learning and
  Systems 2019, MLSys 2019, Stanford, California, USA, March 31 - April 2,
  2019}}, \bibfield{editor}{\bibinfo{person}{Ameet Talwalkar},
  \bibinfo{person}{Virginia Smith}, {and} \bibinfo{person}{Matei Zaharia}}
  (Eds.). \bibinfo{publisher}{mlsys.org}, \bibinfo{pages}{374--388}.
\newblock
\urldef\tempurl%
\url{https://proceedings.mlsys.org/book/271.pdf}
\showURL{%
\tempurl}


\bibitem[\protect\citeauthoryear{Bonawitz, Ivanov, Kreuter, Marcedone, McMahan,
  Patel, Ramage, Segal, and Seth}{Bonawitz et~al\mbox{.}}{2017}]%
        {bonawitz2017practical}
\bibfield{author}{\bibinfo{person}{Kallista~A. Bonawitz},
  \bibinfo{person}{Vladimir Ivanov}, \bibinfo{person}{Ben Kreuter},
  \bibinfo{person}{Antonio Marcedone}, \bibinfo{person}{H.~Brendan McMahan},
  \bibinfo{person}{Sarvar Patel}, \bibinfo{person}{Daniel Ramage},
  \bibinfo{person}{Aaron Segal}, {and} \bibinfo{person}{Karn Seth}.}
  \bibinfo{year}{2017}\natexlab{}.
\newblock \showarticletitle{Practical Secure Aggregation for Privacy-Preserving
  Machine Learning}. In \bibinfo{booktitle}{\emph{Proceedings of the 2017 {ACM}
  {SIGSAC} Conference on Computer and Communications Security, {CCS} 2017,
  Dallas, Texas, USA, October 30 - November 03, 2017}},
  \bibfield{editor}{\bibinfo{person}{Bhavani Thuraisingham},
  \bibinfo{person}{David Evans}, \bibinfo{person}{Tal Malkin}, {and}
  \bibinfo{person}{Dongyan Xu}} (Eds.). \bibinfo{publisher}{{ACM}},
  \bibinfo{pages}{1175--1191}.
\newblock
\urldef\tempurl%
\url{https://doi.org/10.1145/3133956.3133982}
\showDOI{\tempurl}


\bibitem[\protect\citeauthoryear{Brock, Lim, Ritchie, and Weston}{Brock
  et~al\mbox{.}}{2017}]%
        {brock2017freezeout}
\bibfield{author}{\bibinfo{person}{Andrew Brock}, \bibinfo{person}{Theodore
  Lim}, \bibinfo{person}{James~M. Ritchie}, {and} \bibinfo{person}{Nick
  Weston}.} \bibinfo{year}{2017}\natexlab{}.
\newblock \showarticletitle{FreezeOut: Accelerate Training by Progressively
  Freezing Layers}.
\newblock \bibinfo{journal}{\emph{CoRR}}  \bibinfo{volume}{abs/1706.04983}
  (\bibinfo{year}{2017}).
\newblock
\showeprint[arXiv]{1706.04983}
\urldef\tempurl%
\url{http://arxiv.org/abs/1706.04983}
\showURL{%
\tempurl}


\bibitem[\protect\citeauthoryear{Chai, Ali, Zawad, Truex, Anwar, Baracaldo,
  Zhou, Ludwig, Yan, and Cheng}{Chai et~al\mbox{.}}{2020}]%
        {chai2020tifl}
\bibfield{author}{\bibinfo{person}{Zheng Chai}, \bibinfo{person}{Ahsan Ali},
  \bibinfo{person}{Syed Zawad}, \bibinfo{person}{Stacey Truex},
  \bibinfo{person}{Ali Anwar}, \bibinfo{person}{Nathalie Baracaldo},
  \bibinfo{person}{Yi Zhou}, \bibinfo{person}{Heiko Ludwig},
  \bibinfo{person}{Feng Yan}, {and} \bibinfo{person}{Yue Cheng}.}
  \bibinfo{year}{2020}\natexlab{}.
\newblock \showarticletitle{TiFL: {A} Tier-based Federated Learning System}. In
  \bibinfo{booktitle}{\emph{{HPDC} '20: The 29th International Symposium on
  High-Performance Parallel and Distributed Computing, Stockholm, Sweden, June
  23-26, 2020}}, \bibfield{editor}{\bibinfo{person}{Manish Parashar},
  \bibinfo{person}{Vladimir Vlassov}, \bibinfo{person}{David~E. Irwin}, {and}
  \bibinfo{person}{Kathryn Mohror}} (Eds.). \bibinfo{publisher}{{ACM}},
  \bibinfo{pages}{125--136}.
\newblock
\urldef\tempurl%
\url{https://doi.org/10.1145/3369583.3392686}
\showDOI{\tempurl}


\bibitem[\protect\citeauthoryear{Chen, Xu, Wang, Li, Li, Chen, and Zhang}{Chen
  et~al\mbox{.}}{2021}]%
        {chen_apf}
\bibfield{author}{\bibinfo{person}{Chen Chen}, \bibinfo{person}{Hong Xu},
  \bibinfo{person}{Wei Wang}, \bibinfo{person}{Baochun Li}, \bibinfo{person}{Bo
  Li}, \bibinfo{person}{Li Chen}, {and} \bibinfo{person}{Gong Zhang}.}
  \bibinfo{year}{2021}\natexlab{}.
\newblock \showarticletitle{Communication-Efficient Federated Learning with
  Adaptive Parameter Freezing}. In \bibinfo{booktitle}{\emph{41st {IEEE}
  International Conference on Distributed Computing Systems, {ICDCS} 2021,
  Washington DC, USA, July 7-10, 2021}}. \bibinfo{publisher}{{IEEE}},
  \bibinfo{pages}{1--11}.
\newblock
\urldef\tempurl%
\url{https://doi.org/10.1109/ICDCS51616.2021.00010}
\showDOI{\tempurl}


\bibitem[\protect\citeauthoryear{Costan and Devadas}{Costan and
  Devadas}{2016}]%
        {costan2016intel}
\bibfield{author}{\bibinfo{person}{Victor Costan} {and}
  \bibinfo{person}{Srinivas Devadas}.} \bibinfo{year}{2016}\natexlab{}.
\newblock \showarticletitle{Intel {SGX} Explained}.
\newblock \bibinfo{journal}{\emph{{IACR} Cryptol. ePrint Arch.}}
  (\bibinfo{year}{2016}), \bibinfo{pages}{86}.
\newblock
\urldef\tempurl%
\url{http://eprint.iacr.org/2016/086}
\showURL{%
\tempurl}


\bibitem[\protect\citeauthoryear{Cox, Galjaard, Ghiassi, Birke, and Chen}{Cox
  et~al\mbox{.}}{2021}]%
        {cox2021masa}
\bibfield{author}{\bibinfo{person}{Bart Cox}, \bibinfo{person}{Jeroen
  Galjaard}, \bibinfo{person}{Amirmasoud Ghiassi}, \bibinfo{person}{Robert
  Birke}, {and} \bibinfo{person}{Lydia~Y. Chen}.}
  \bibinfo{year}{2021}\natexlab{}.
\newblock \showarticletitle{Masa: Responsive Multi-DNN Inference on the Edge}.
  In \bibinfo{booktitle}{\emph{19th {IEEE} International Conference on
  Pervasive Computing and Communications, PerCom 2021, Kassel, Germany, March
  22-26, 2021}}. \bibinfo{publisher}{{IEEE}}, \bibinfo{pages}{1--10}.
\newblock
\urldef\tempurl%
\url{https://doi.org/10.1109/PERCOM50583.2021.9439111}
\showDOI{\tempurl}


\bibitem[\protect\citeauthoryear{Dong, Zeng, Gu, and Guo}{Dong
  et~al\mbox{.}}{2020}]%
        {DBLP:conf/mass/DongZ0G20}
\bibfield{author}{\bibinfo{person}{Shifu Dong}, \bibinfo{person}{Deze Zeng},
  \bibinfo{person}{Lin Gu}, {and} \bibinfo{person}{Song Guo}.}
  \bibinfo{year}{2020}\natexlab{}.
\newblock \showarticletitle{Offloading Federated Learning Task to Edge
  Computing with Trust Execution Environment}. In
  \bibinfo{booktitle}{\emph{17th {IEEE} International Conference on Mobile Ad
  Hoc and Sensor Systems, {MASS} 2020, Delhi, India, December 10-13, 2020}}.
  \bibinfo{publisher}{{IEEE}}, \bibinfo{pages}{491--496}.
\newblock
\urldef\tempurl%
\url{https://doi.org/10.1109/MASS50613.2020.00066}
\showDOI{\tempurl}


\bibitem[\protect\citeauthoryear{Graham}{Graham}{1969}]%
        {graham1969bounds}
\bibfield{author}{\bibinfo{person}{Ronald~L. Graham}.}
  \bibinfo{year}{1969}\natexlab{}.
\newblock \showarticletitle{Bounds on Multiprocessing Timing Anomalies}.
\newblock \bibinfo{journal}{\emph{{SIAM} Journal of Applied Mathematics}}
  \bibinfo{volume}{17}, \bibinfo{number}{2} (\bibinfo{year}{1969}),
  \bibinfo{pages}{416--429}.
\newblock
\urldef\tempurl%
\url{https://doi.org/10.1137/0117039}
\showURL{%
\tempurl}


\bibitem[\protect\citeauthoryear{Hsieh, Phanishayee, Mutlu, and Gibbons}{Hsieh
  et~al\mbox{.}}{2020}]%
        {hsieh2020non}
\bibfield{author}{\bibinfo{person}{Kevin Hsieh}, \bibinfo{person}{Amar
  Phanishayee}, \bibinfo{person}{Onur Mutlu}, {and} \bibinfo{person}{Phillip~B.
  Gibbons}.} \bibinfo{year}{2020}\natexlab{}.
\newblock \showarticletitle{The Non-IID Data Quagmire of Decentralized Machine
  Learning}. In \bibinfo{booktitle}{\emph{Proceedings of the 37th International
  Conference on Machine Learning, {ICML} 2020, 13-18 July 2020, Virtual Event}}
  \emph{(\bibinfo{series}{Proceedings of Machine Learning Research})},
  Vol.~\bibinfo{volume}{119}. \bibinfo{publisher}{{PMLR}},
  \bibinfo{pages}{4387--4398}.
\newblock
\urldef\tempurl%
\url{http://proceedings.mlr.press/v119/hsieh20a.html}
\showURL{%
\tempurl}


\bibitem[\protect\citeauthoryear{Ji, Chen, Zhao, Chen, Wei, and Yu}{Ji
  et~al\mbox{.}}{2021}]%
        {ji2021computation}
\bibfield{author}{\bibinfo{person}{Zhongming Ji}, \bibinfo{person}{Li Chen},
  \bibinfo{person}{Nan Zhao}, \bibinfo{person}{Yunfei Chen},
  \bibinfo{person}{Guo Wei}, {and} \bibinfo{person}{F.~Richard Yu}.}
  \bibinfo{year}{2021}\natexlab{}.
\newblock \showarticletitle{Computation Offloading for Edge-Assisted Federated
  Learning}.
\newblock \bibinfo{journal}{\emph{{IEEE} Trans. Veh. Technol.}}
  \bibinfo{volume}{70}, \bibinfo{number}{9} (\bibinfo{year}{2021}),
  \bibinfo{pages}{9330--9344}.
\newblock
\urldef\tempurl%
\url{https://doi.org/10.1109/TVT.2021.3098022}
\showDOI{\tempurl}


\bibitem[\protect\citeauthoryear{Kairouz, McMahan, Avent, Bellet, Bennis,
  Bhagoji, Bonawitz, Charles, Cormode, Cummings, D'Oliveira, Eichner, Rouayheb,
  Evans, Gardner, Garrett, Gasc{\'{o}}n, Ghazi, Gibbons, Gruteser, Harchaoui,
  He, He, Huo, Hutchinson, Hsu, Jaggi, Javidi, Joshi, Khodak,
  Kone{\v{c}}n{\'y}, Korolova, Koushanfar, Koyejo, Lepoint, Liu, Mittal, Mohri,
  Nock, {\"{O}}zg{\"{u}}r, Pagh, Qi, Ramage, Raskar, Raykova, Song, Song,
  Stich, Sun, Suresh, Tram{\`{e}}r, Vepakomma, Wang, Xiong, Xu, Yang, Yu, Yu,
  and Zhao}{Kairouz et~al\mbox{.}}{2021}]%
        {kairouz2021advances}
\bibfield{author}{\bibinfo{person}{Peter Kairouz}, \bibinfo{person}{H.~Brendan
  McMahan}, \bibinfo{person}{Brendan Avent}, \bibinfo{person}{Aur{\'{e}}lien
  Bellet}, \bibinfo{person}{Mehdi Bennis}, \bibinfo{person}{Arjun~Nitin
  Bhagoji}, \bibinfo{person}{Kallista~A. Bonawitz}, \bibinfo{person}{Zachary
  Charles}, \bibinfo{person}{Graham Cormode}, \bibinfo{person}{Rachel
  Cummings}, \bibinfo{person}{Rafael G.~L. D'Oliveira}, \bibinfo{person}{Hubert
  Eichner}, \bibinfo{person}{Salim~El Rouayheb}, \bibinfo{person}{David Evans},
  \bibinfo{person}{Josh Gardner}, \bibinfo{person}{Zachary Garrett},
  \bibinfo{person}{Adri{\`{a}} Gasc{\'{o}}n}, \bibinfo{person}{Badih Ghazi},
  \bibinfo{person}{Phillip~B. Gibbons}, \bibinfo{person}{Marco Gruteser},
  \bibinfo{person}{Za{\"{\i}}d Harchaoui}, \bibinfo{person}{Chaoyang He},
  \bibinfo{person}{Lie He}, \bibinfo{person}{Zhouyuan Huo},
  \bibinfo{person}{Ben Hutchinson}, \bibinfo{person}{Justin Hsu},
  \bibinfo{person}{Martin Jaggi}, \bibinfo{person}{Tara Javidi},
  \bibinfo{person}{Gauri Joshi}, \bibinfo{person}{Mikhail Khodak},
  \bibinfo{person}{Jakub Kone{\v{c}}n{\'y}}, \bibinfo{person}{Aleksandra
  Korolova}, \bibinfo{person}{Farinaz Koushanfar}, \bibinfo{person}{Sanmi
  Koyejo}, \bibinfo{person}{Tancr{\`{e}}de Lepoint}, \bibinfo{person}{Yang
  Liu}, \bibinfo{person}{Prateek Mittal}, \bibinfo{person}{Mehryar Mohri},
  \bibinfo{person}{Richard Nock}, \bibinfo{person}{Ayfer {\"{O}}zg{\"{u}}r},
  \bibinfo{person}{Rasmus Pagh}, \bibinfo{person}{Hang Qi},
  \bibinfo{person}{Daniel Ramage}, \bibinfo{person}{Ramesh Raskar},
  \bibinfo{person}{Mariana Raykova}, \bibinfo{person}{Dawn Song},
  \bibinfo{person}{Weikang Song}, \bibinfo{person}{Sebastian~U. Stich},
  \bibinfo{person}{Ziteng Sun}, \bibinfo{person}{Ananda~Theertha Suresh},
  \bibinfo{person}{Florian Tram{\`{e}}r}, \bibinfo{person}{Praneeth Vepakomma},
  \bibinfo{person}{Jianyu Wang}, \bibinfo{person}{Li Xiong},
  \bibinfo{person}{Zheng Xu}, \bibinfo{person}{Qiang Yang},
  \bibinfo{person}{Felix~X. Yu}, \bibinfo{person}{Han Yu}, {and}
  \bibinfo{person}{Sen Zhao}.} \bibinfo{year}{2021}\natexlab{}.
\newblock \showarticletitle{Advances and Open Problems in Federated Learning}.
\newblock \bibinfo{journal}{\emph{Found. Trends Mach. Learn.}}
  \bibinfo{volume}{14}, \bibinfo{number}{1-2} (\bibinfo{year}{2021}),
  \bibinfo{pages}{1--210}.
\newblock
\urldef\tempurl%
\url{https://doi.org/10.1561/2200000083}
\showDOI{\tempurl}


\bibitem[\protect\citeauthoryear{Karimireddy, Kale, Mohri, Reddi, Stich, and
  Suresh}{Karimireddy et~al\mbox{.}}{2020}]%
        {karimireddy2020scaffold}
\bibfield{author}{\bibinfo{person}{Sai~Praneeth Karimireddy},
  \bibinfo{person}{Satyen Kale}, \bibinfo{person}{Mehryar Mohri},
  \bibinfo{person}{Sashank~J. Reddi}, \bibinfo{person}{Sebastian~U. Stich},
  {and} \bibinfo{person}{Ananda~Theertha Suresh}.}
  \bibinfo{year}{2020}\natexlab{}.
\newblock \showarticletitle{{SCAFFOLD:} Stochastic Controlled Averaging for
  Federated Learning}. In \bibinfo{booktitle}{\emph{Proceedings of the 37th
  International Conference on Machine Learning, {ICML} 2020, 13-18 July 2020,
  Virtual Event}} \emph{(\bibinfo{series}{Proceedings of Machine Learning
  Research})}, Vol.~\bibinfo{volume}{119}. \bibinfo{publisher}{{PMLR}},
  \bibinfo{pages}{5132--5143}.
\newblock
\urldef\tempurl%
\url{http://proceedings.mlr.press/v119/karimireddy20a.html}
\showURL{%
\tempurl}


\bibitem[\protect\citeauthoryear{Krizhevsky and Hinton}{Krizhevsky and
  Hinton}{2009}]%
        {krizhevsky2009learning}
\bibfield{author}{\bibinfo{person}{Alex Krizhevsky} {and}
  \bibinfo{person}{Geoffrey Hinton}.} \bibinfo{year}{2009}\natexlab{}.
\newblock \bibinfo{booktitle}{\emph{Learning multiple layers of features from
  tiny images}}.
\newblock \bibinfo{type}{{T}echnical {R}eport}~0.
  \bibinfo{institution}{University of Toronto}, \bibinfo{address}{Toronto,
  Ontario}.
\newblock


\bibitem[\protect\citeauthoryear{Lawler, Lenstra, Kan, and Shmoys}{Lawler
  et~al\mbox{.}}{1993}]%
        {lawler89sequencing}
\bibfield{author}{\bibinfo{person}{Eugene~L. Lawler},
  \bibinfo{person}{Jan~Karel Lenstra}, \bibinfo{person}{Alexander H. G.~Rinnooy
  Kan}, {and} \bibinfo{person}{David~B. Shmoys}.}
  \bibinfo{year}{1993}\natexlab{}.
\newblock \showarticletitle{Chapter 9 Sequencing and scheduling: Algorithms and
  complexity}.
\newblock In \bibinfo{booktitle}{\emph{Logistics of Production and Inventory}},
  \bibfield{editor}{\bibinfo{person}{Stephen~C. Graves},
  \bibinfo{person}{Alexander H. G.~Rinnooy Kan}, {and}
  \bibinfo{person}{Paul~Herbert Zipkin}} (Eds.). \bibinfo{series}{Handbooks in
  Operations Research and Management Science}, Vol.~\bibinfo{volume}{4}.
  \bibinfo{publisher}{North-Holland}, \bibinfo{pages}{445--522}.
\newblock
\urldef\tempurl%
\url{https://doi.org/10.1016/s0927-0507(05)80189-6}
\showDOI{\tempurl}


\bibitem[\protect\citeauthoryear{LeCun, Bottou, Bengio, and Haffner}{LeCun
  et~al\mbox{.}}{1998}]%
        {lecun1998gradient}
\bibfield{author}{\bibinfo{person}{Yann LeCun}, \bibinfo{person}{L{\'{e}}on
  Bottou}, \bibinfo{person}{Yoshua Bengio}, {and} \bibinfo{person}{Patrick
  Haffner}.} \bibinfo{year}{1998}\natexlab{}.
\newblock \showarticletitle{Gradient-based learning applied to document
  recognition}.
\newblock \bibinfo{journal}{\emph{Proc. {IEEE}}} \bibinfo{volume}{86},
  \bibinfo{number}{11} (\bibinfo{year}{1998}), \bibinfo{pages}{2278--2324}.
\newblock
\urldef\tempurl%
\url{https://doi.org/10.1109/5.726791}
\showDOI{\tempurl}


\bibitem[\protect\citeauthoryear{Lee, Ko, and Pack}{Lee et~al\mbox{.}}{2022}]%
        {lee2021adaptive}
\bibfield{author}{\bibinfo{person}{Jaewook Lee}, \bibinfo{person}{Haneul Ko},
  {and} \bibinfo{person}{Sangheon Pack}.} \bibinfo{year}{2022}\natexlab{}.
\newblock \showarticletitle{Adaptive Deadline Determination for Mobile Device
  Selection in Federated Learning}.
\newblock \bibinfo{journal}{\emph{{IEEE} Trans. Veh. Technol.}}
  \bibinfo{volume}{71}, \bibinfo{number}{3} (\bibinfo{year}{2022}),
  \bibinfo{pages}{3367--3371}.
\newblock
\urldef\tempurl%
\url{https://doi.org/10.1109/TVT.2021.3136308}
\showDOI{\tempurl}


\bibitem[\protect\citeauthoryear{Li, Xiong, Guo, Wang, and Xu}{Li
  et~al\mbox{.}}{2019}]%
        {li2019smartpc}
\bibfield{author}{\bibinfo{person}{Li Li}, \bibinfo{person}{Haoyi Xiong},
  \bibinfo{person}{Zhishan Guo}, \bibinfo{person}{Jun Wang}, {and}
  \bibinfo{person}{Cheng{-}Zhong Xu}.} \bibinfo{year}{2019}\natexlab{}.
\newblock \showarticletitle{SmartPC: Hierarchical Pace Control in Real-Time
  Federated Learning System}. In \bibinfo{booktitle}{\emph{{IEEE} Real-Time
  Systems Symposium, {RTSS} 2019, Hong Kong, SAR, China, December 3-6, 2019}}.
  \bibinfo{publisher}{{IEEE}}, \bibinfo{pages}{406--418}.
\newblock
\urldef\tempurl%
\url{https://doi.org/10.1109/RTSS46320.2019.00043}
\showDOI{\tempurl}


\bibitem[\protect\citeauthoryear{Li, Sahu, Zaheer, Sanjabi, Talwalkar, and
  Smith}{Li et~al\mbox{.}}{2020}]%
        {li2018federated}
\bibfield{author}{\bibinfo{person}{Tian Li}, \bibinfo{person}{Anit~Kumar Sahu},
  \bibinfo{person}{Manzil Zaheer}, \bibinfo{person}{Maziar Sanjabi},
  \bibinfo{person}{Ameet Talwalkar}, {and} \bibinfo{person}{Virginia Smith}.}
  \bibinfo{year}{2020}\natexlab{}.
\newblock \showarticletitle{Federated Optimization in Heterogeneous Networks}.
  In \bibinfo{booktitle}{\emph{Proceedings of Machine Learning and Systems
  2020, MLSys 2020, Austin, Texas, USA, March 2-4, 2020}},
  \bibfield{editor}{\bibinfo{person}{Inderjit~S. Dhillon},
  \bibinfo{person}{Dimitris~S. Papailiopoulos}, {and} \bibinfo{person}{Vivienne
  Sze}} (Eds.). \bibinfo{publisher}{mlsys.org}, \bibinfo{pages}{429--450}.
\newblock
\urldef\tempurl%
\url{https://proceedings.mlsys.org/book/316.pdf}
\showURL{%
\tempurl}


\bibitem[\protect\citeauthoryear{McMahan, Moore, Ramage, Hampson, and
  y~Arcas}{McMahan et~al\mbox{.}}{2017}]%
        {mcmahan2017communication}
\bibfield{author}{\bibinfo{person}{Brendan McMahan}, \bibinfo{person}{Eider
  Moore}, \bibinfo{person}{Daniel Ramage}, \bibinfo{person}{Seth Hampson},
  {and} \bibinfo{person}{Blaise~Ag{\"{u}}era y Arcas}.}
  \bibinfo{year}{2017}\natexlab{}.
\newblock \showarticletitle{Communication-Efficient Learning of Deep Networks
  from Decentralized Data}. In \bibinfo{booktitle}{\emph{Proceedings of the
  20th International Conference on Artificial Intelligence and Statistics,
  {AISTATS} 2017, 20-22 April 2017, Fort Lauderdale, Florida, {USA}}}
  \emph{(\bibinfo{series}{Proceedings of Machine Learning Research})},
  \bibfield{editor}{\bibinfo{person}{Aarti Singh} {and}
  \bibinfo{person}{Xiaojin~(Jerry) Zhu}} (Eds.), Vol.~\bibinfo{volume}{54}.
  \bibinfo{publisher}{{PMLR}}, \bibinfo{pages}{1273--1282}.
\newblock
\urldef\tempurl%
\url{http://proceedings.mlr.press/v54/mcmahan17a.html}
\showURL{%
\tempurl}


\bibitem[\protect\citeauthoryear{McMahan, Moore, Ramage, and y~Arcas}{McMahan
  et~al\mbox{.}}{2016}]%
        {mcmahan2016federated}
\bibfield{author}{\bibinfo{person}{H.~Brendan McMahan}, \bibinfo{person}{Eider
  Moore}, \bibinfo{person}{Daniel Ramage}, {and}
  \bibinfo{person}{Blaise~Ag{\"{u}}era y Arcas}.}
  \bibinfo{year}{2016}\natexlab{}.
\newblock \showarticletitle{Federated Learning of Deep Networks using Model
  Averaging}.
\newblock \bibinfo{journal}{\emph{CoRR}}  \bibinfo{volume}{abs/1602.05629}
  (\bibinfo{year}{2016}).
\newblock
\showeprint[arXiv]{1602.05629}
\urldef\tempurl%
\url{http://arxiv.org/abs/1602.05629}
\showURL{%
\tempurl}


\bibitem[\protect\citeauthoryear{Nishio and Yonetani}{Nishio and
  Yonetani}{2019}]%
        {nishio2019client}
\bibfield{author}{\bibinfo{person}{Takayuki Nishio} {and} \bibinfo{person}{Ryo
  Yonetani}.} \bibinfo{year}{2019}\natexlab{}.
\newblock \showarticletitle{Client Selection for Federated Learning with
  Heterogeneous Resources in Mobile Edge}. In \bibinfo{booktitle}{\emph{2019
  {IEEE} International Conference on Communications, {ICC} 2019, Shanghai,
  China, May 20-24, 2019}}. \bibinfo{publisher}{{IEEE}}, \bibinfo{pages}{1--7}.
\newblock
\urldef\tempurl%
\url{https://doi.org/10.1109/ICC.2019.8761315}
\showDOI{\tempurl}


\bibitem[\protect\citeauthoryear{Oquab, Bottou, Laptev, and Sivic}{Oquab
  et~al\mbox{.}}{2014}]%
        {oquab2014learning}
\bibfield{author}{\bibinfo{person}{Maxime Oquab}, \bibinfo{person}{L{\'{e}}on
  Bottou}, \bibinfo{person}{Ivan Laptev}, {and} \bibinfo{person}{Josef Sivic}.}
  \bibinfo{year}{2014}\natexlab{}.
\newblock \showarticletitle{Learning and Transferring Mid-level Image
  Representations Using Convolutional Neural Networks}. In
  \bibinfo{booktitle}{\emph{2014 {IEEE} Conference on Computer Vision and
  Pattern Recognition, {CVPR} 2014, Columbus, Ohio, USA, June 23-28, 2014}}.
  \bibinfo{publisher}{{IEEE} Computer Society}, \bibinfo{pages}{1717--1724}.
\newblock
\urldef\tempurl%
\url{https://doi.org/10.1109/CVPR.2014.222}
\showDOI{\tempurl}


\bibitem[\protect\citeauthoryear{Orm{\'{a}}ndi, Heged{\"{u}}s, and
  Jelasity}{Orm{\'{a}}ndi et~al\mbox{.}}{2013}]%
        {ormandi2013gossip}
\bibfield{author}{\bibinfo{person}{R{\'{o}}bert Orm{\'{a}}ndi},
  \bibinfo{person}{Istv{\'{a}}n Heged{\"{u}}s}, {and}
  \bibinfo{person}{M{\'{a}}rk Jelasity}.} \bibinfo{year}{2013}\natexlab{}.
\newblock \showarticletitle{Gossip learning with linear models on fully
  distributed data}.
\newblock \bibinfo{journal}{\emph{Concurr. Comput. Pract. Exp.}}
  \bibinfo{volume}{25}, \bibinfo{number}{4} (\bibinfo{year}{2013}),
  \bibinfo{pages}{556--571}.
\newblock
\urldef\tempurl%
\url{https://doi.org/10.1002/cpe.2858}
\showDOI{\tempurl}


\bibitem[\protect\citeauthoryear{Park, Han, Choi, and Moon}{Park
  et~al\mbox{.}}{2021}]%
        {park2021sageflow}
\bibfield{author}{\bibinfo{person}{Jungwuk Park}, \bibinfo{person}{Dong{-}Jun
  Han}, \bibinfo{person}{Minseok Choi}, {and} \bibinfo{person}{Jaekyun Moon}.}
  \bibinfo{year}{2021}\natexlab{}.
\newblock \showarticletitle{Sageflow: Robust Federated Learning against Both
  Stragglers and Adversaries}. In \bibinfo{booktitle}{\emph{Advances in Neural
  Information Processing Systems 34: Annual Conference on Neural Information
  Processing Systems 2021, NeurIPS 2021, December 6-14, 2021, virtual}},
  \bibfield{editor}{\bibinfo{person}{Marc'Aurelio Ranzato},
  \bibinfo{person}{Alina Beygelzimer}, \bibinfo{person}{Yann~N. Dauphin},
  \bibinfo{person}{Percy Liang}, {and} \bibinfo{person}{Jennifer~Wortman
  Vaughan}} (Eds.). \bibinfo{pages}{840--851}.
\newblock
\urldef\tempurl%
\url{https://proceedings.neurips.cc/paper/2021/hash/076a8133735eb5d7552dc195b125a454-Abstract.html}
\showURL{%
\tempurl}


\bibitem[\protect\citeauthoryear{Rubner, Tomasi, and Guibas}{Rubner
  et~al\mbox{.}}{1998}]%
        {rubner1998metric}
\bibfield{author}{\bibinfo{person}{Yossi Rubner}, \bibinfo{person}{Carlo
  Tomasi}, {and} \bibinfo{person}{Leonidas~J. Guibas}.}
  \bibinfo{year}{1998}\natexlab{}.
\newblock \showarticletitle{A Metric for Distributions with Applications to
  Image Databases}. In \bibinfo{booktitle}{\emph{Proceedings of the Sixth
  International Conference on Computer Vision (ICCV-98), Bombay, India, January
  4-7, 1998}}. \bibinfo{publisher}{{IEEE} Computer Society},
  \bibinfo{pages}{59--66}.
\newblock
\urldef\tempurl%
\url{https://doi.org/10.1109/ICCV.1998.710701}
\showDOI{\tempurl}


\bibitem[\protect\citeauthoryear{Sun, Chen, Wang, Liu, and Liu}{Sun
  et~al\mbox{.}}{2016}]%
        {DBLP:conf/aaai/SunCWLL16}
\bibfield{author}{\bibinfo{person}{Shizhao Sun}, \bibinfo{person}{Wei Chen},
  \bibinfo{person}{Liwei Wang}, \bibinfo{person}{Xiaoguang Liu}, {and}
  \bibinfo{person}{Tie{-}Yan Liu}.} \bibinfo{year}{2016}\natexlab{}.
\newblock \showarticletitle{On the Depth of Deep Neural Networks: {A}
  Theoretical View}. In \bibinfo{booktitle}{\emph{Proceedings of the Thirtieth
  {AAAI} Conference on Artificial Intelligence, February 12-17, 2016, Phoenix,
  Arizona, {USA}}}, \bibfield{editor}{\bibinfo{person}{Dale Schuurmans} {and}
  \bibinfo{person}{Michael~P. Wellman}} (Eds.). \bibinfo{publisher}{{AAAI}
  Press}, \bibinfo{pages}{2066--2072}.
\newblock
\urldef\tempurl%
\url{http://www.aaai.org/ocs/index.php/AAAI/AAAI16/paper/view/12073}
\showURL{%
\tempurl}


\bibitem[\protect\citeauthoryear{Tang}{Tang}{2013}]%
        {tang2013deep}
\bibfield{author}{\bibinfo{person}{Yichuan Tang}.}
  \bibinfo{year}{2013}\natexlab{}.
\newblock \showarticletitle{Deep Learning using Support Vector Machines}.
\newblock \bibinfo{journal}{\emph{CoRR}}  \bibinfo{volume}{abs/1306.0239}
  (\bibinfo{year}{2013}).
\newblock
\showeprint[arXiv]{1306.0239}
\urldef\tempurl%
\url{http://arxiv.org/abs/1306.0239}
\showURL{%
\tempurl}


\bibitem[\protect\citeauthoryear{Tsai, Porter, and Vij}{Tsai
  et~al\mbox{.}}{2017}]%
        {tsai2017graphene}
\bibfield{author}{\bibinfo{person}{Chia{-}che Tsai}, \bibinfo{person}{Donald~E.
  Porter}, {and} \bibinfo{person}{Mona Vij}.} \bibinfo{year}{2017}\natexlab{}.
\newblock \showarticletitle{Graphene-SGX: {A} Practical Library {OS} for
  Unmodified Applications on {SGX}}. In \bibinfo{booktitle}{\emph{2017 {USENIX}
  Annual Technical Conference, {USENIX} {ATC} 2017, Santa Clara, California,
  USA, July 12-14, 2017}}, \bibfield{editor}{\bibinfo{person}{Dilma~Da Silva}
  {and} \bibinfo{person}{Bryan Ford}} (Eds.). \bibinfo{publisher}{{USENIX}
  Association}, \bibinfo{pages}{645--658}.
\newblock
\urldef\tempurl%
\url{https://www.usenix.org/conference/atc17/technical-sessions/presentation/tsai}
\showURL{%
\tempurl}


\bibitem[\protect\citeauthoryear{Veit and Belongie}{Veit and Belongie}{2018}]%
        {veit2018convolutional}
\bibfield{author}{\bibinfo{person}{Andreas Veit} {and}
  \bibinfo{person}{Serge~J. Belongie}.} \bibinfo{year}{2018}\natexlab{}.
\newblock \showarticletitle{Convolutional Networks with Adaptive Inference
  Graphs}. In \bibinfo{booktitle}{\emph{Computer Vision - {ECCV} 2018 - 15th
  European Conference, Munich, Germany, September 8-14, 2018, Proceedings, Part
  {I}}} \emph{(\bibinfo{series}{Lecture Notes in Computer Science})},
  \bibfield{editor}{\bibinfo{person}{Vittorio Ferrari},
  \bibinfo{person}{Martial Hebert}, \bibinfo{person}{Cristian Sminchisescu},
  {and} \bibinfo{person}{Yair Weiss}} (Eds.), Vol.~\bibinfo{volume}{11205}.
  \bibinfo{publisher}{Springer}, \bibinfo{pages}{3--18}.
\newblock
\urldef\tempurl%
\url{https://doi.org/10.1007/978-3-030-01246-5\_1}
\showDOI{\tempurl}


\bibitem[\protect\citeauthoryear{Wang, Liu, Liang, Joshi, and Poor}{Wang
  et~al\mbox{.}}{2020}]%
        {wang2020tackling}
\bibfield{author}{\bibinfo{person}{Jianyu Wang}, \bibinfo{person}{Qinghua Liu},
  \bibinfo{person}{Hao Liang}, \bibinfo{person}{Gauri Joshi}, {and}
  \bibinfo{person}{H.~Vincent Poor}.} \bibinfo{year}{2020}\natexlab{}.
\newblock \showarticletitle{Tackling the Objective Inconsistency Problem in
  Heterogeneous Federated Optimization}. In \bibinfo{booktitle}{\emph{Advances
  in Neural Information Processing Systems 33: Annual Conference on Neural
  Information Processing Systems 2020, NeurIPS 2020, December 6-12, 2020,
  virtual}}, \bibfield{editor}{\bibinfo{person}{Hugo Larochelle},
  \bibinfo{person}{Marc'Aurelio Ranzato}, \bibinfo{person}{Raia Hadsell},
  \bibinfo{person}{Maria{-}Florina Balcan}, {and} \bibinfo{person}{Hsuan{-}Tien
  Lin}} (Eds.). \bibinfo{pages}{7611--7623}.
\newblock
\urldef\tempurl%
\url{https://proceedings.neurips.cc/paper/2020/hash/564127c03caab942e503ee6f810f54fd-Abstract.html}
\showURL{%
\tempurl}


\bibitem[\protect\citeauthoryear{Wu, Brooks, Chen, Chen, Choudhury, Dukhan,
  Hazelwood, Isaac, Jia, Jia, Leyvand, Lu, Lu, Qiao, Reagen, Spisak, Sun,
  Tulloch, Vajda, Wang, Wang, Wasti, Wu, Xian, Yoo, and Zhang}{Wu
  et~al\mbox{.}}{2019}]%
        {wu2019machine}
\bibfield{author}{\bibinfo{person}{Carole{-}Jean Wu}, \bibinfo{person}{David
  Brooks}, \bibinfo{person}{Kevin Chen}, \bibinfo{person}{Douglas Chen},
  \bibinfo{person}{Sy Choudhury}, \bibinfo{person}{Marat Dukhan},
  \bibinfo{person}{Kim~M. Hazelwood}, \bibinfo{person}{Eldad Isaac},
  \bibinfo{person}{Yangqing Jia}, \bibinfo{person}{Bill Jia},
  \bibinfo{person}{Tommer Leyvand}, \bibinfo{person}{Hao Lu},
  \bibinfo{person}{Yang Lu}, \bibinfo{person}{Lin Qiao},
  \bibinfo{person}{Brandon Reagen}, \bibinfo{person}{Joe Spisak},
  \bibinfo{person}{Fei Sun}, \bibinfo{person}{Andrew Tulloch},
  \bibinfo{person}{Peter Vajda}, \bibinfo{person}{Xiaodong Wang},
  \bibinfo{person}{Yanghan Wang}, \bibinfo{person}{Bram Wasti},
  \bibinfo{person}{Yiming Wu}, \bibinfo{person}{Ran Xian},
  \bibinfo{person}{Sungjoo Yoo}, {and} \bibinfo{person}{Peizhao Zhang}.}
  \bibinfo{year}{2019}\natexlab{}.
\newblock \showarticletitle{Machine Learning at Facebook: Understanding
  Inference at the Edge}. In \bibinfo{booktitle}{\emph{25th {IEEE}
  International Symposium on High Performance Computer Architecture, {HPCA}
  2019, Washington, DC, USA, February 16-20, 2019}}.
  \bibinfo{publisher}{{IEEE}}, \bibinfo{pages}{331--344}.
\newblock
\urldef\tempurl%
\url{https://doi.org/10.1109/HPCA.2019.00048}
\showDOI{\tempurl}


\bibitem[\protect\citeauthoryear{Wu, Ullah, Harvey, Kilpatrick, Spence, and
  Varghese}{Wu et~al\mbox{.}}{2021}]%
        {wu2021fedadapt}
\bibfield{author}{\bibinfo{person}{Di Wu}, \bibinfo{person}{Rehmat Ullah},
  \bibinfo{person}{Paul Harvey}, \bibinfo{person}{Peter Kilpatrick},
  \bibinfo{person}{Ivor T.~A. Spence}, {and} \bibinfo{person}{Blesson
  Varghese}.} \bibinfo{year}{2021}\natexlab{}.
\newblock \showarticletitle{FedAdapt: Adaptive Offloading for IoT Devices in
  Federated Learning}.
\newblock \bibinfo{journal}{\emph{CoRR}}  \bibinfo{volume}{abs/2107.04271}
  (\bibinfo{year}{2021}).
\newblock
\showeprint[arXiv]{2107.04271}
\urldef\tempurl%
\url{https://arxiv.org/abs/2107.04271}
\showURL{%
\tempurl}


\bibitem[\protect\citeauthoryear{Wu, Nagarajan, Kumar, Rennie, Davis, Grauman,
  and Feris}{Wu et~al\mbox{.}}{2018}]%
        {wu2018blockdrop}
\bibfield{author}{\bibinfo{person}{Zuxuan Wu}, \bibinfo{person}{Tushar
  Nagarajan}, \bibinfo{person}{Abhishek Kumar}, \bibinfo{person}{Steven
  Rennie}, \bibinfo{person}{Larry~S. Davis}, \bibinfo{person}{Kristen Grauman},
  {and} \bibinfo{person}{Rog{\'{e}}rio~Schmidt Feris}.}
  \bibinfo{year}{2018}\natexlab{}.
\newblock \showarticletitle{BlockDrop: Dynamic Inference Paths in Residual
  Networks}. In \bibinfo{booktitle}{\emph{2018 {IEEE} Conference on Computer
  Vision and Pattern Recognition, {CVPR} 2018, Salt Lake City, Utah, USA, June
  18-22, 2018}}. \bibinfo{publisher}{Computer Vision Foundation / {IEEE}
  Computer Society}, \bibinfo{pages}{8817--8826}.
\newblock
\urldef\tempurl%
\url{https://doi.org/10.1109/CVPR.2018.00919}
\showDOI{\tempurl}


\bibitem[\protect\citeauthoryear{Xiao, Mudiyanselage, Ji, Hu, and Pan}{Xiao
  et~al\mbox{.}}{2019}]%
        {xiao2019fast}
\bibfield{author}{\bibinfo{person}{Xueli Xiao}, \bibinfo{person}{Thosini~Bamunu
  Mudiyanselage}, \bibinfo{person}{Chunyan Ji}, \bibinfo{person}{Jie Hu}, {and}
  \bibinfo{person}{Yi Pan}.} \bibinfo{year}{2019}\natexlab{}.
\newblock \showarticletitle{Fast Deep Learning Training through Intelligently
  Freezing Layers}. In \bibinfo{booktitle}{\emph{2019 International Conference
  on Internet of Things (iThings) and {IEEE} Green Computing and Communications
  (GreenCom) and {IEEE} Cyber, Physical and Social Computing (CPSCom) and
  {IEEE} Smart Data (SmartData), iThings/GreenCom/CPSCom/SmartData 2019,
  Atlanta, Georgia, USA, July 14-17, 2019}}. \bibinfo{publisher}{{IEEE}},
  \bibinfo{pages}{1225--1232}.
\newblock
\urldef\tempurl%
\url{https://doi.org/10.1109/iThings/GreenCom/CPSCom/SmartData.2019.00205}
\showDOI{\tempurl}


\bibitem[\protect\citeauthoryear{Zhang, Guo, Ma, Wang, Xu, and Wu}{Zhang
  et~al\mbox{.}}{2021}]%
        {zhang2021parameterized}
\bibfield{author}{\bibinfo{person}{Jie Zhang}, \bibinfo{person}{Song Guo},
  \bibinfo{person}{Xiaosong Ma}, \bibinfo{person}{Haozhao Wang},
  \bibinfo{person}{Wenchao Xu}, {and} \bibinfo{person}{Feijie Wu}.}
  \bibinfo{year}{2021}\natexlab{}.
\newblock \showarticletitle{Parameterized Knowledge Transfer for Personalized
  Federated Learning}. In \bibinfo{booktitle}{\emph{Advances in Neural
  Information Processing Systems 34: Annual Conference on Neural Information
  Processing Systems 2021, NeurIPS 2021, December 6-14, 2021, virtual}},
  \bibfield{editor}{\bibinfo{person}{Marc'Aurelio Ranzato},
  \bibinfo{person}{Alina Beygelzimer}, \bibinfo{person}{Yann~N. Dauphin},
  \bibinfo{person}{Percy Liang}, {and} \bibinfo{person}{Jennifer~Wortman
  Vaughan}} (Eds.). \bibinfo{pages}{10092--10104}.
\newblock
\urldef\tempurl%
\url{https://proceedings.neurips.cc/paper/2021/hash/5383c7318a3158b9bc261d0b6996f7c2-Abstract.html}
\showURL{%
\tempurl}


\bibitem[\protect\citeauthoryear{Zhang and Sabuncu}{Zhang and Sabuncu}{2018}]%
        {zhang2018generalized}
\bibfield{author}{\bibinfo{person}{Zhilu Zhang} {and} \bibinfo{person}{Mert~R.
  Sabuncu}.} \bibinfo{year}{2018}\natexlab{}.
\newblock \showarticletitle{Generalized Cross Entropy Loss for Training Deep
  Neural Networks with Noisy Labels}. In \bibinfo{booktitle}{\emph{Advances in
  Neural Information Processing Systems 31: Annual Conference on Neural
  Information Processing Systems 2018, NeurIPS 2018, December 3-8, 2018,
  Montr{\'{e}}al, Canada}}, \bibfield{editor}{\bibinfo{person}{Samy Bengio},
  \bibinfo{person}{Hanna~M. Wallach}, \bibinfo{person}{Hugo Larochelle},
  \bibinfo{person}{Kristen Grauman}, \bibinfo{person}{Nicol{\`{o}}
  Cesa{-}Bianchi}, {and} \bibinfo{person}{Roman Garnett}} (Eds.).
  \bibinfo{pages}{8792--8802}.
\newblock
\urldef\tempurl%
\url{https://proceedings.neurips.cc/paper/2018/hash/f2925f97bc13ad2852a7a551802feea0-Abstract.html}
\showURL{%
\tempurl}


\bibitem[\protect\citeauthoryear{Zhao, Li, Lai, Suda, Civin, and Chandra}{Zhao
  et~al\mbox{.}}{2018}]%
        {zhao2018federated}
\bibfield{author}{\bibinfo{person}{Yue Zhao}, \bibinfo{person}{Meng Li},
  \bibinfo{person}{Liangzhen Lai}, \bibinfo{person}{Naveen Suda},
  \bibinfo{person}{Damon Civin}, {and} \bibinfo{person}{Vikas Chandra}.}
  \bibinfo{year}{2018}\natexlab{}.
\newblock \showarticletitle{Federated Learning with Non-IID Data}.
\newblock \bibinfo{journal}{\emph{CoRR}}  \bibinfo{volume}{abs/1806.00582}
  (\bibinfo{year}{2018}).
\newblock
\showeprint[arXiv]{1806.00582}
\urldef\tempurl%
\url{http://arxiv.org/abs/1806.00582}
\showURL{%
\tempurl}


\end{thebibliography}

\end{document}